\documentclass[letterpaper, 10 pt, conference]{ieeeconf}
\IEEEoverridecommandlockouts
\pdfoutput = 1 
\usepackage{amsmath,amssymb,amsfonts}
\usepackage{algorithmic,algorithm}
\usepackage{amsmath,graphicx,amsfonts,amssymb}
\usepackage{mathrsfs}
\usepackage{graphicx}
\usepackage{textcomp}
\usepackage{caption}
\usepackage{subcaption}
\usepackage{cite}
\usepackage{siunitx}
\usepackage{hyperref}
\usepackage[flushleft]{threeparttable}
\usepackage{url}
\usepackage{color}
\title{\LARGE \bf
Are We Ready for Unmanned Surface Vehicles in Inland Waterways? 

The USVInland Multisensor Dataset and Benchmark
}

\author{Yuwei Cheng$^{1}$, Mengxin Jiang$^{2}$, Jiannan Zhu$^{3}$, Yimin Liu$^{4}$
\thanks{$^{1}$Yuwei Cheng is with the Department of Electronic Engineering, Tsinghua University, Beijing, 100084 China, and also with the ORCA-TECH, Shaanxi, 710075 China {\tt\small chengyw18@mails.tsinghua.edu.cn}}
\thanks{$^{2}$Mengxin Jiang is with the Department of Automation, Tsinghua University, Beijing, 10084 China, and also with the ORCA-TECH, Shaanxi, 710075 China {\tt\small jiangmx16@mails.tsinghua.edu.cn}}
\thanks{$^{3}$Jiannan Zhu is with the School of Marine Science and Technology, Northwestern Polytechnical University, Shaanxi, 710109 China, and also with the ORCA-TECH, Shaanxi, 710075 China {\tt\small jacknyzhu@orca-tech.com.cn}}
\thanks{$^{4}$Yimin Liu, as the corresponding author, is with the Department of Electronic Engineering, Tsinghua University, Beijing, 100084 China {\tt\small yiminliu@tsinghua.edu.cn}}
}

\begin{document}

\maketitle
\thispagestyle{empty}
\pagestyle{empty}

\begin{abstract}
Unmanned surface vehicles (USVs) have great value with their ability to execute hazardous and time-consuming missions over water surfaces. Recently, USVs for inland waterways have attracted increasing attention for their potential application in autonomous monitoring, transportation, and cleaning. However, unlike sailing in open water, the challenges posed by scenes of inland waterways, such as the complex distribution of obstacles, the global positioning system (GPS) signal denial environment, the reflection of bank-side structures, and the fog over the water surface, all impede USV application in inland waterways. To address these problems and stimulate relevant research, we introduce USVInland, a multisensor dataset for USVs in inland waterways. The collection of USVInland spans a trajectory of more than 26 km in diverse real-world scenes of inland waterways using various modalities, including lidar, stereo cameras, millimeter-wave radar, GPS, and inertial measurement units (IMUs). Based on the requirements and challenges in the perception and navigation of USVs for inland waterways, we build benchmarks for simultaneous localization and mapping (SLAM), stereo matching, and water segmentation. We evaluate common algorithms for the above tasks to determine the influence of unique inland waterway scenes on algorithm performance. Our dataset and the development tools are available online at \href{https://www.orca-tech.cn/datasets.html}{https://www.orca-tech.cn/datasets.html}.

% the performance of the well-used algorithms still not meets the requirement of considering the variations in waterway structures and weather conditions.

~\
\end{abstract}


\section{Introduction}
\begin{figure}
\centering
\vspace{0.1in}
\begin{subfigure}{0.23\textwidth}
\centering
\includegraphics[width=\textwidth]{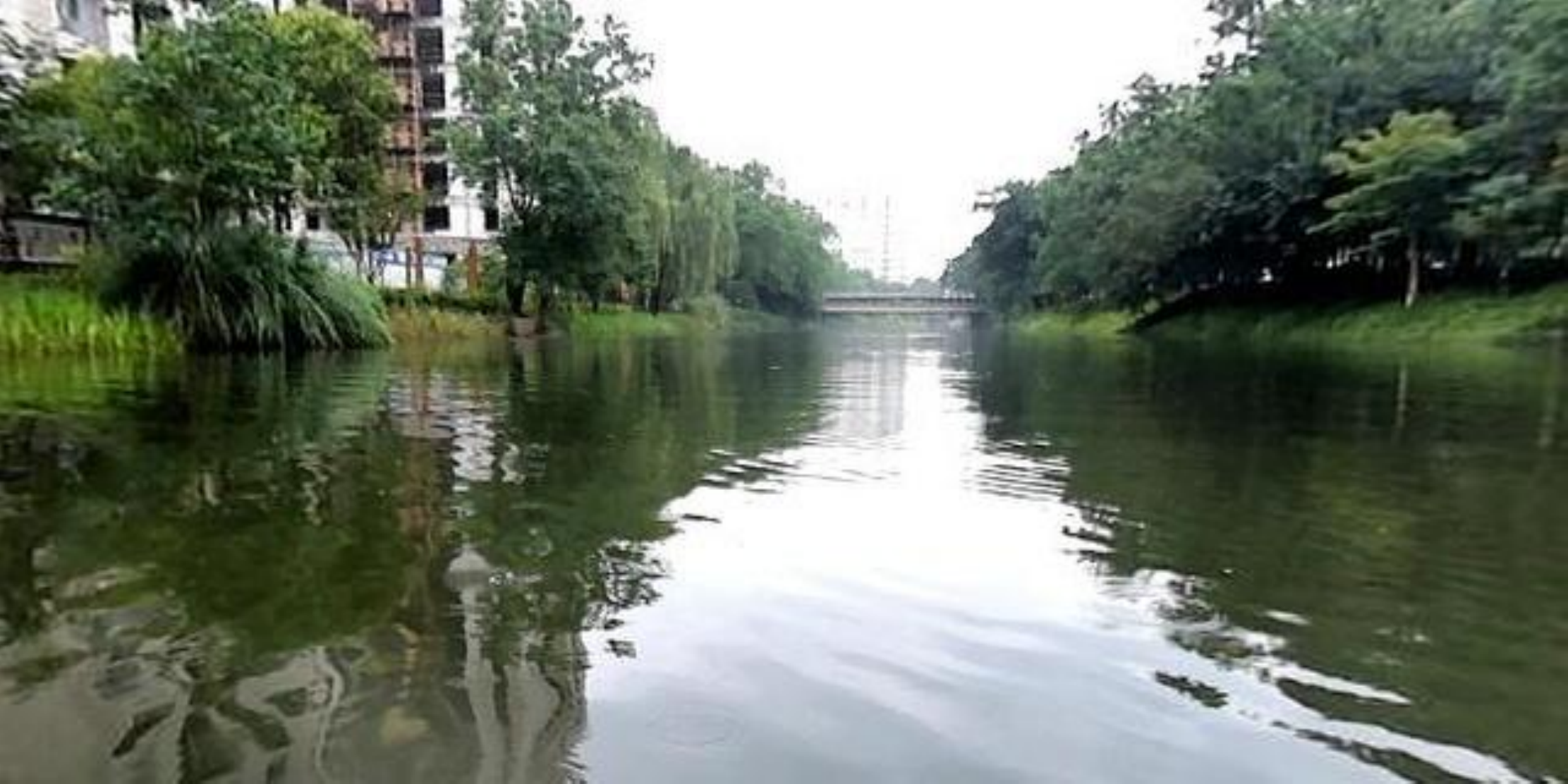} 
\caption{Normal}
\label{fig:Normal}
\end{subfigure}
\begin{subfigure}{0.23\textwidth}
\centering
\includegraphics[width=\textwidth]{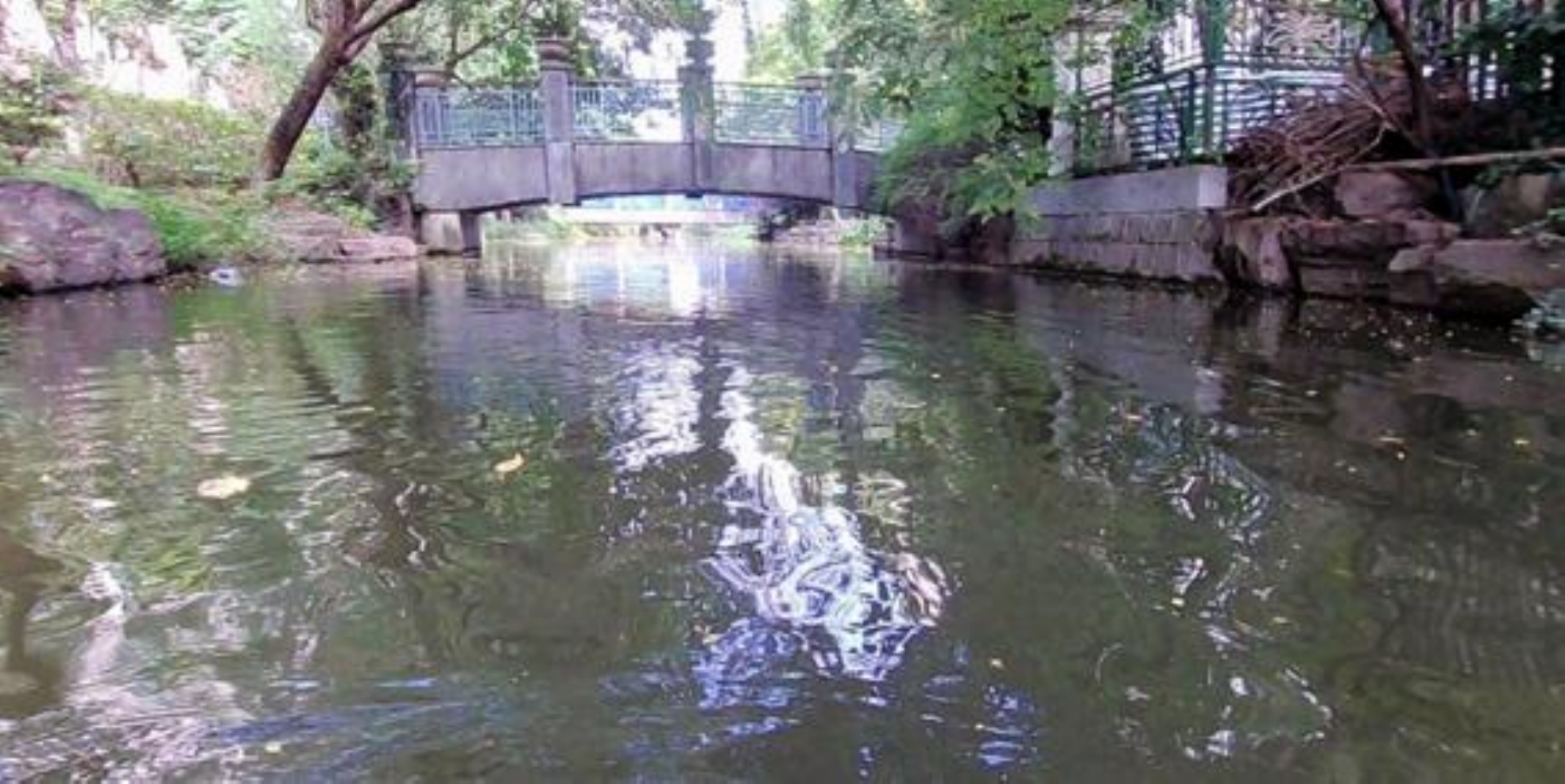}
\caption{Narrow}
\label{fig:Narrow}
\end{subfigure}

\begin{subfigure}{0.23\textwidth}
\centering
\includegraphics[width=\textwidth]{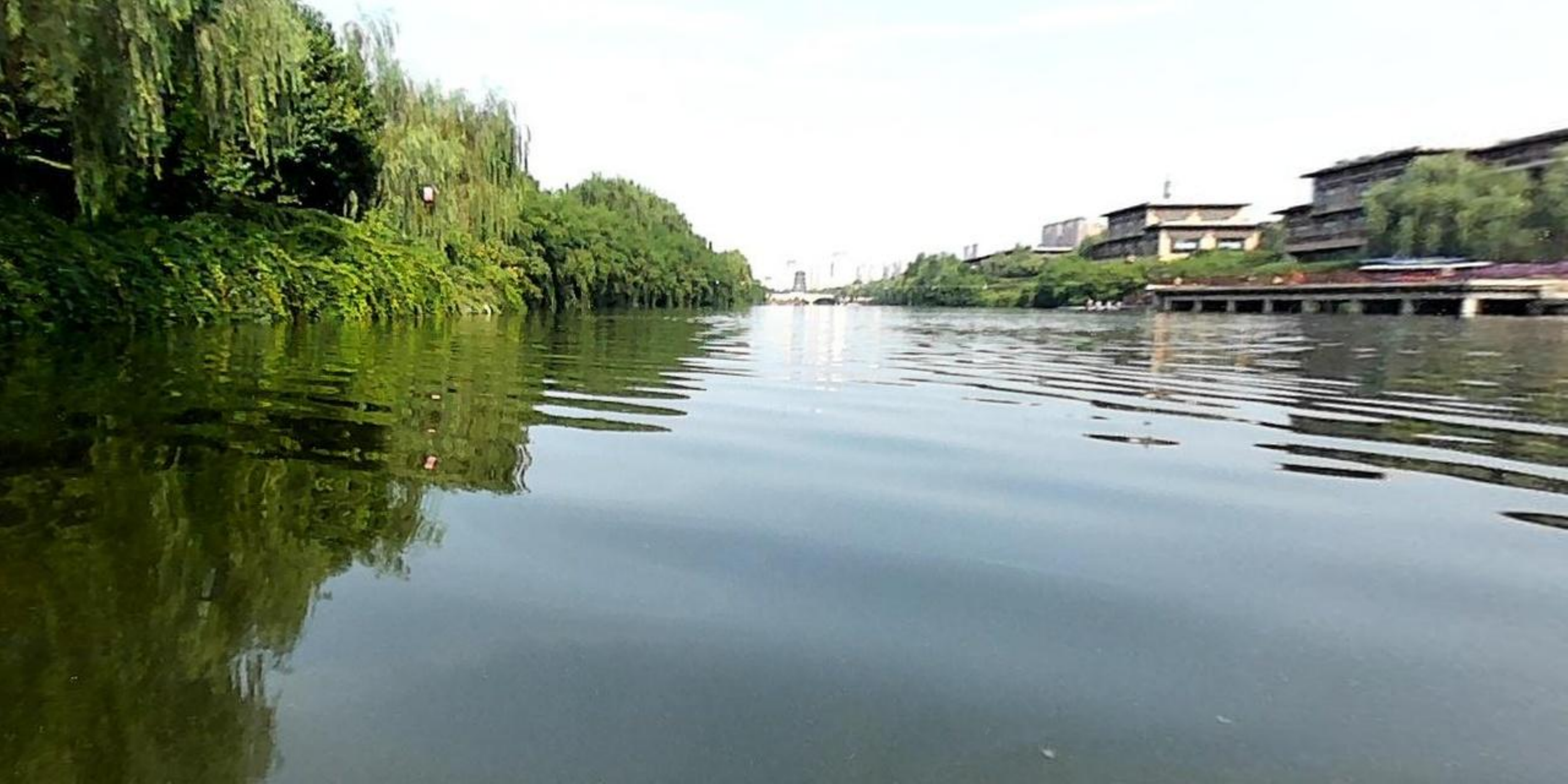}
\caption{Wide}
\label{fig:Wide}
\end{subfigure}
\begin{subfigure}{0.23\textwidth}
\centering
\includegraphics[width=\textwidth]{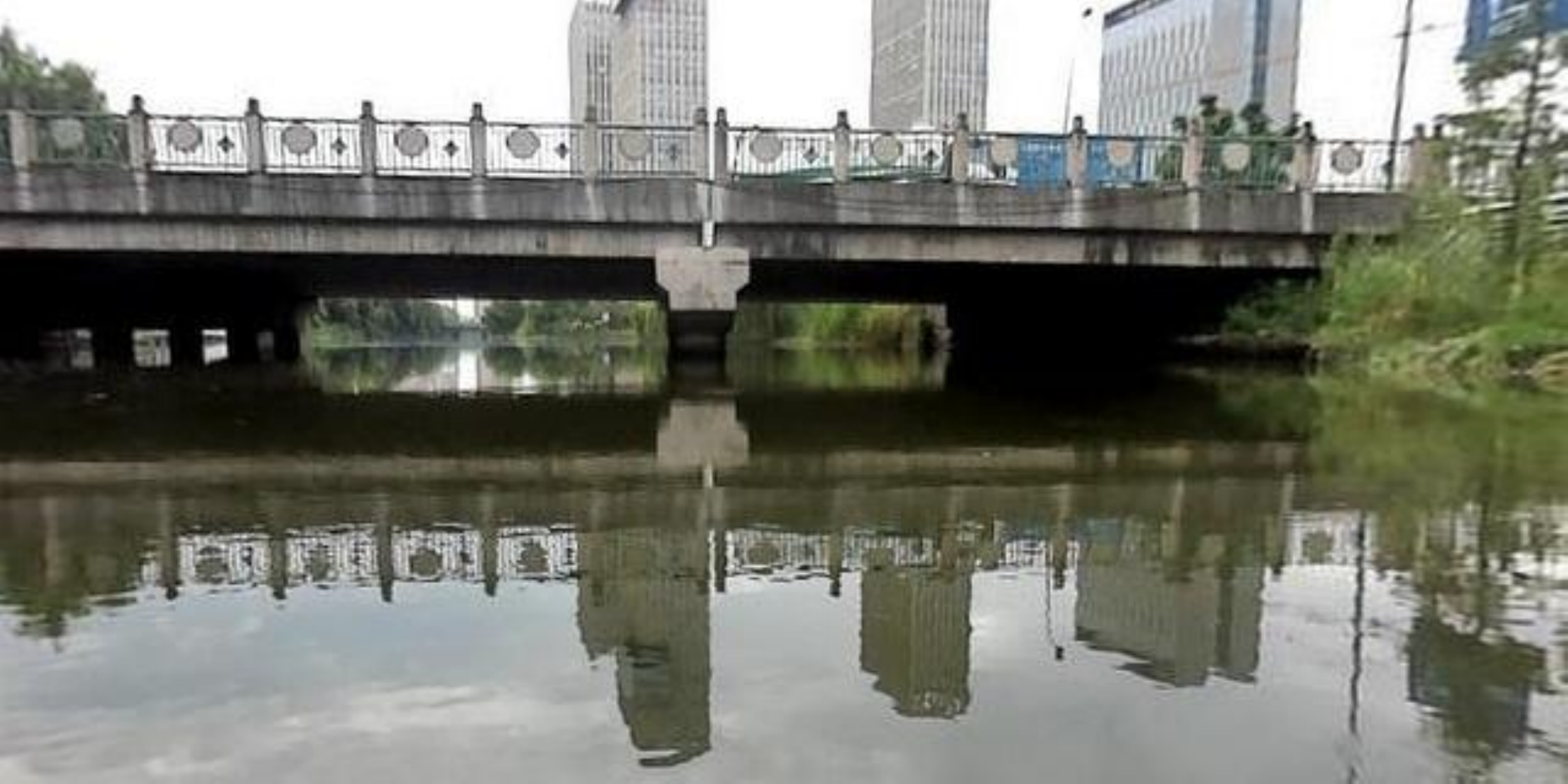}
\caption{Complex}
\label{fig:Complex}
\end{subfigure}

\caption{A variety of inland waterway landscapes and structures.}
\label{fig:inland waterway}
\end{figure}

In recent years, unmanned surface vehicles (USVs), also known as autonomous surface vehicles (ASVs), have gained increasing prominence driven by their ability to perform hazardous and time-consuming missions \cite{han2019coastal}. 
The strong demands from commercial, scientific and environmental communities accelerate the development of USV applications, such as hydrographic surveying and charting, marine resource explorations, water quality monitoring, and floating waste removal \cite{peng2017development,pastore2010improving,madeo2020low,ruangpayoongsak2017floating}.
Compared to coastal and marine USVs, USVs for inland waterways are more closely related to human life and have a large potential value, such as being the core of building autonomous transportation systems in inland waterways \cite{wang2019roboat}.
Despite this, USVs for inland waterways have not been fully developed mainly due to the challenges of the variable and complex real-world driving scenes of inland waterways \cite{wang2019roboat}, as shown in Fig.~\ref{fig:inland waterway}. 
In narrow inland waterways, global positioning system (GPS) signals are sometimes attenuated due to the occlusion of riparian vegetation, bridges, and urban settlements \cite{kriechbaumer2015quantitative}. In this case, to achieve reliable navigation in inland waterways, accurate and real-time localization relies on the estimation of the vehicle's relative location to the surrounding environment \cite{fuentes2015visual}. In addition, it is essential to keep the vehicle at a safe distance from the bank and other obstacles. Thus, similar to autonomous driving on roads, the tasks of simultaneous localization and mapping (SLAM), stereo matching, and water segmentation based on sensors such as lidar, stereo cameras, and millimeter-wave radar are introduced to support the localization, perception, and navigation of USVs for inland waterways.

However, in inland waterways, lidar data noise caused by fog over the water surface and strong light reflection would compromise the lidar system \cite{bijelic2018benchmark,wang2020roboat}. 
In terms of the visual system, surface fog can reduce visibility and obscure scenes. The images might be overexposed due to sunlight reflection. 
In addition, the reflection of bankside objects and the rippling caused by raindrops also interfere with the visual system. Besides, in inland waterways, USVs pose changes more frequently due to the low surface friction and water waves, leading to difficulties in matching landmarks in adjacent frames for some SLAM algorithms. Common strategies for autonomous driving and marine USVs are infeasible for USVs in inland waterways.
% The distinctions between road landscapes and waterway landscapes, the complex environment, and more prominent change of poses on the wavy water surface, all bring new challenges to USVs for the inland waterway.

%%%%%%%%%%%%%%%%%%%%%%%%%%%%%%%%%%%%%%%% figure_inland_waterway %%%%%%%%%%%%%%%%%%%%%%%%%%%%%%%%%%%%%
%%%%%%%%%%%%%%%%%%%%%%%%%%%%%%%%%%%%%%%% figure_inland_waterway %%%%%%%%%%%%%%%%%%%%%%%%%%%%%%%%%%%%%

%%%%%%%%%%%%%%%%%%%%%%%%%%%%%%%%%%%%%%%% figure_inland_waterway %%%%%%%%%%%%%%%%%%%%%%%%%%%%%%%%%%%%%
%%%%%%%%%%%%%%%%%%%%%%%%%%%%%%%%%%%%%%%% figure_inland_waterway %%%%%%%%%%%%%%%%%%%%%%%%%%%%%%%%%%%%%

Publicly available datasets can encourage breakthroughs and enable a fair comparison between different algorithms. For autonomous driving, advances have been achieved through exploration on some widespread datasets, such as the KITTI \cite{geiger2012we}, RobotCar \cite{maddern20171}, 
and Cityscapes datasets \cite{cordts2016cityscapes}. Datasets such as the Stanford Drone Dataset \cite{robicquet2016learning} and the Okutama-Action dataset \cite{barekatain2017okutama} also promote the development of unmanned aerial vehicles (UAVs). For USVs, datasets for certain tasks in marine and coastal areas, including object detection \cite{zhang2015vais,prasad2017video} and water segmentation \cite{taipalmaa2019high,bovcon2019mastr1325}, have been published. However, to the best of our knowledge, there is no public standard dataset aimed at USVs for inland waterways or a dataset for USVs containing data from various sensors that makes multiple tasks and sensor fusion methods possible.

To fill this gap and help research USVs, we introduce USVInland, a multisensor dataset of USVs for inland waterways collected under a variety of weather conditions. Typical sensors used in autonomous driving, including stereo cameras, a lidar system, GPS antennas, and inertial measurement units (IMUs), are mounted on our acquisition platform. Three millimeter-wave radars, which are less expensive and more robust to weather conditions than lidar systems, are also used to capture data of the environment around the vehicle. All the sensors are synchronized, and the calibration results are provided in our dataset.

Inspired by the KITTI visual benchmark \cite{geiger2012we}, with reference to the tasks for the autonomous vehicle, we introduce SLAM, stereo matching, and water segmentation tasks into our dataset and build corresponding benchmarks. Experiments on different tasks are carried out on our dataset based on typical algorithms. The results show that the performance of commonly used algorithms would decrease due to the characteristics of inland waterway scenes.
% to meet the requirement of USVs perception and navigation in inland waterways.
% The result shows that these algorithms may not meet the requirements of USVs' perception and navigation in inland waterways.
%The result shows that many algorithms well-used in autonomous driving have worse performance when applied to scenes in inland waterways.

The USVInland dataset along with the calibration data are available online on our website. To make the dataset more accessible to researchers, we provide development tools on the website for loading and visualizing synchronized data from multiple sensors. The introduction of the file formats can also be found on the website.

To summarize, this paper mainly contributes to the following aspects:
\begin{itemize}
\item
We present the USVInland dataset, the first dataset for USVs in inland waterways, containing data from multiple sensors, including a lidar system, stereo cameras, millimeter-wave radar, GPS antennas, and IMUs.

\item 
Our dataset is collected in real-world scenes of inland waterways at different times, covering a variety of waterway landscapes and weather conditions. We provide accurate timestamps for the synchronization of different sensors as well as intrinsic and extrinsic sensor calibration results.

\item
Aiming at the requirements and challenges in the perception and navigation of USVs for inland waterways, we introduce the tasks of SLAM, stereo matching, and water segmentation. We build corresponding benchmarks and evaluate common algorithms for different tasks on our benchmarks.

%  It also contains a large amount of data collected under harsh conditions.
% which would spur studies towards USVs for inland waterways and enable comparability for different perception strategies.
\end{itemize}

\section{Related Work}
\subsection{USVs for the Inland Waterway}
With advances in autonomous vehicles, the potential applications of USVs in inland waterways have been proven to be achievable and partly converted into practical applications in fields of monitoring, cleaning, and transportation. Shojaei \textit{et al.} \cite{shojaei2018proof} present their study on using USVs for structural health monitoring, which is based on detecting cracks in buildings near water bodies by visual information. The solar-powered Lake Wivenhoe USV was designed to monitor the water quality properties and greenhouse gas emissions \cite{dunbabin2009autonomous}. The Water Environmental Mobile Observer (WeMo) developed by Madeo \textit{et al.} \cite{madeo2020low} collects data on water depth in addition to water quality.
For water body cleaning, the USVs presented in \cite{akib2019unmanned} and \cite{ruangpayoongsak2017floating} were developed to collect floating waste above the water surface in inland waterways. In addition, the Roboat project \cite{wang2019roboat,wang2020roboat} aims at autonomous transportation and dynamic floating infrastructure construction in urban waterways. Their current boat includes a complete design of localization, perception, planning, and control to support autonomous transportation tasks.

\subsection{SLAM}
SLAM is an essential capability for the self-navigation of self-driving vehicles in an unknown environment. Based on lidar and cameras, constantly emerging approaches such as LOAM \cite{zhang2014loam}, Cartographer \cite{hess2016real}, and LeGO-LOAM \cite{shan2018lego} for lidar SLAM, as well as ORB-SLAM \cite{mur2015orb} and VINS \cite{qin2018vins} for visual SLAM, have been proven to be valuable for navigation systems of self-driving vehicles and have been transferred into practical use.
% In addition, SLAM approaches based on the radar as described in  \cite{vivet2013localization,schuster2016landmark,cen2018precise}, have attracted researches and are proven to be more robust to harsh and diverse conditions.

In recent years, 77 GHz automotive radar technology has rapidly developed. The resolution of radar point clouds becomes higher, providing richer information on the surrounding environment. Meanwhile, some researchers pay attention to SLAM based on millimeter-wave radar \cite{meyer2019automotive,barnes2019oxford, vivet2013localization,schuster2016landmark}. It is worth investigating the application value of millimeter-wave radar on USVs because radar is more robust to harsh weather conditions, such as fog and rain than lidar systems and cameras and it is not influenced by lighting conditions.
%as a low-cost alternative to Lidar in the perception of autonomous driving, 

% As the GPS signal denial areas cause trouble for USVs navigation, SLAM for USVs becomes particularly important.
For USVs, the SLAM algorithm has been used to overcome the difficulties of navigation in the GPS signal denial area. Han and Kim \cite{han2013navigation} proposed a method to estimate the location of the vehicle under bridges using a landmark-based SLAM framework based on lidar point clouds. Wang \textit{et al.} \cite{wang2020roboat} applied LIO-SAM \cite{shan2020lio}, a tightly-coupled Lidar inertial odometry, on their urban waterway transportation USV, Roboat II. Kriechbaumer \textit{et al.} \cite{kriechbaumer2015quantitative} focused on visual odometry used in USV localization, and they evaluate both the feature-based and appearance-based visual odometry algorithms to develop a river monitoring USV. Han \textit{et al.} \cite{han2019coastal} proposed C-SLAM for USV navigation and mapping in coastal water, which is based on the coastline features extracted from a marine radar image.

\subsection{Stereo Matching}
Stereo matching, which is used to determine the disparity map between an image pair taken from the same scene for depth information \cite{hong2004segment}, is always an essential task for the perception of self-driving vehicles.
%has been intensively investigated for several decades.%[copy]
Classical methods such as Semi-Global Matching (SGM) \cite{hirschmuller2007stereo} and recently proposed methods such as those from \cite{vzbontar2016stereo,luo2016efficient,yang2018segstereo} enable increasingly accurate depth information of the surrounding environment for self-driving vehicles.
%disparity map and 

% Hirschmuller and Heiko propose the classical method based on Semi-Global Matching(SGM)\cite{hirschmuller2007stereo}. 
% The convolution neural network(CNN) based algorithms firstly proposed in\cite{vzbontar2016stereo} are then explored on stereo matching and achieve strong performance\cite{luo2016efficient}\cite{shaked2017improved}. Unsupervised stereo matching\cite{jason2016back}\cite{meister2017unflow} and semantic-guided algorithms\cite{guney2015displets}\cite{yang2018segstereo} are also developed and 
% promoted to increase the accuracy of disparity maps.

For USVs, stereo matching is widely used for obstacle detection. For example, Wang and Wei \cite{wang2013stereovision} developed a marine USV system for obstacle detection and tracking by building a grid map generated from stereo reconstructed 3D points.
Based on the idea of combining semantic segmentation and stereo matching, the approach introduced in  \cite{bovcon2018obstacle} performs well in segmenting obstacles in marine environments by aligning image pairs to enforce segmentation consistency.

\subsection{Water Segmentation}
Studies on detecting water area have been performed to enable safe navigation for both marine and inland waterways USVs. Most of the current detection methods are based on visual information.
Zhan \textit{et al.} \cite{zhan2017effective} propose their method in which the waterline is fitted based on the extracted potential boundary points using random sample consensus (RANSAC). Another method described in \cite{zou2020novel} first extracts line segments in an image based on gradient value and then generates the onshore line segment pool to distinguish the shore area and water area.
% \cite{jianhong2013adaptive} applies color segmentation and Hough Transform to detect the shorelines. 

However, the above methods may fail when the shoreline is extremely irregular in a complex inland water environment or when there is interference over the water surface. Recently, water area detection based on detecting the water boundaries has been treated as a water segmentation task based on deep learning.
% \cite{zhan2019autonomous} segments water, sky, and shore area by initial feature and color based segmentation and U-net combined with conditional random field (CRF) refining. 
Zhan \textit{et al.} \cite{zhan2019autonomous} proposed an online learning approach to segment water, sky, and shore areas in unknown
navigation environments using a convolutional neural network (CNN).
Bovcon and Kristan \cite{bovcon2020water} 
proposed a water-obstacle separation and refinement network that outperforms current semantic segmentation models.
Based on the semantic segmentation model, water area detection is not limited to areas with straight boundaries.
% and is more likely to be applied on the perception and navigation systems of USVs.

\section{Dataset}  

\subsection{Platform}
\begin{figure}
    \vspace{0.08in}
    \centering
    \begin{subfigure}{0.45\textwidth}
        \includegraphics[width=\textwidth]{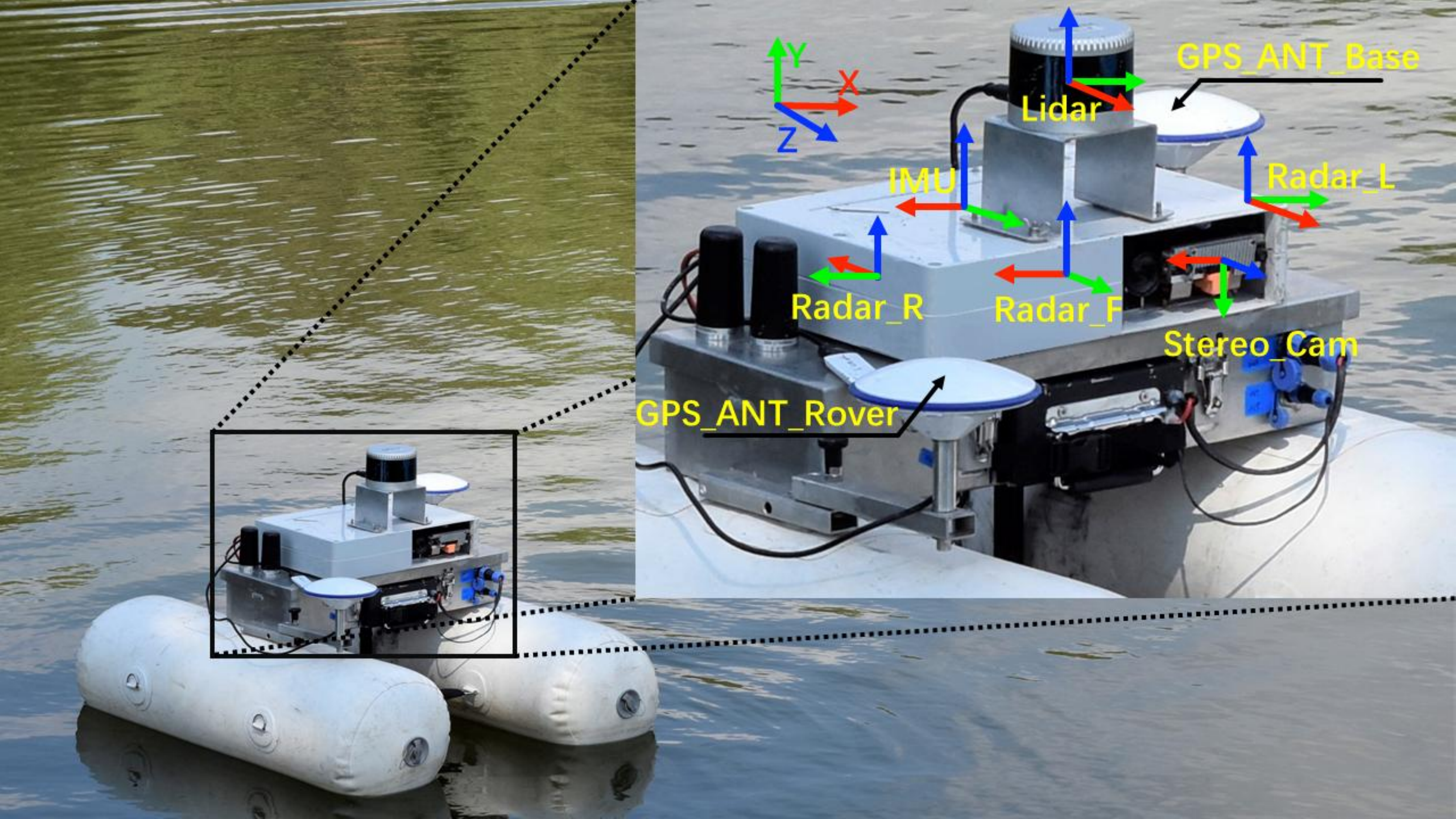}
        \label{fig:platform photo}
    \end{subfigure}
    \caption{The inflatable boat equipped with multiple sensors. 
    The directions of the sensors are marked in different colors (X-red, Y-green, Z-blue).}
    \label{fig:platform}
\end{figure}

Our acquisition platform consists of an inflatable surface vehicle with multiple sensors that enables benchmarks for various tasks. The platform is equipped with a 16-beam lidar system, a stereo color camera, three millimeter-wave radars, and an inertial navigation system (INS). The locations of the sensors on our platform are shown in Fig.~\ref{fig:platform}. A lidar system that captures accurate 3D properties around our boat is mounted on the top of the platform. The stereo camera is placed in front of the boat facing forward. As the horizontal field of view (HFoV) of the millimeter-wave radar is limited, we mounted three radars at the front and two sides of the acquisition setup to collect more comprehensive radar data around the vehicle. 
Two GPS antennas are mounted on two nonadjacent corners of the boat connected with the base and rover module inside of the GPS.
One of them is set at the real-time kinematic (RTK) mode receiving signals from the base station to provide up to centimeter-level accuracy. The other module is connected with the rover module to obtain accurate absolute heading information by the moving base support in the module firmware \cite{Ublox:ZED-F9P}.
% To record more accurate trajectories, the localization system receives RTK correction signal to provide up to centimeter-level accuracy. 
% connected with the base and rover modules inside of the GPS to get accurate absolute heading information by moving base support in the module firmware. 

%%%%%%%%%%%%%%%%%%%%%%%%%%%%%%%%%%%%%%%% figure_platform %%%%%%%%%%%%%%%%%%%%%%%%%%%%%%%%%%%%%
%%%%%%%%%%%%%%%%%%%%%%%%%%%%%%%%%%%%%%%% figure_platform %%%%%%%%%%%%%%%%%%%%%%%%%%%%%%%%%%%%%
%%%%%%%%%%%%%%%%%%%%%%%%%%%%%%%%%%%%%%%% figure_platform %%%%%%%%%%%%%%%%%%%%%%%%%%%%%%%%%%%%%
%%%%%%%%%%%%%%%%%%%%%%%%%%%%%%%%%%%%%%%% figure_platform %%%%%%%%%%%%%%%%%%%%%%%%%%%%%%%%%%%%%

Detailed information about the sensors of the platform is described as follows:

\textbf{1 × Lidar}: 905 nm, 16 beams, HFoV $\ang{360}$ × Vertical Field of View (VFoV) $\ang{32}$, 10 Hz, Max Range 100 m, Range Resolution 2 cm, Horizontal Angular Resolution $\ang{0.18}$

\textbf{1 × Stereo Camera with 1 × IMU}: RGB: 640 × 400/1280 × 800, 20Hz, HFoV$\ang{95}$ × VFoV$\ang{50}$ × Diagonal Field of View (DFoV) $\ang{112}$, Global Shutter, 80 mm  Baseline, Hardware Synchronized; IMU: 200 Hz, 6-axis

\textbf{3 × Radar}: 10 Hz, HFoV$\ang{94}$ × VFoV$\ang{50}$ (-6dB), 
Range Resolution 0.045m, 
Maximum Unambiguous Range 18.08m, 
Maximum Radial Velocity 3.96m/s,
Radial Velocity Resolution 0.25m/s

\textbf{1 × Inertial Navigation System}: IMU 50Hz, 6-axis, 
2 × GPS: U-Blox ZED-F9, 5Hz, Receive RTK Signal, Include Moving Base Support

% \textcolor{blue}{
More information including the make and model of each sensor can be found on our website.
% }

\subsection{Data Collection}
\begin{figure}
\vspace{0.08in}
\centering
\begin{subfigure}{0.15\textwidth}
\centering
\includegraphics[width=\textwidth]{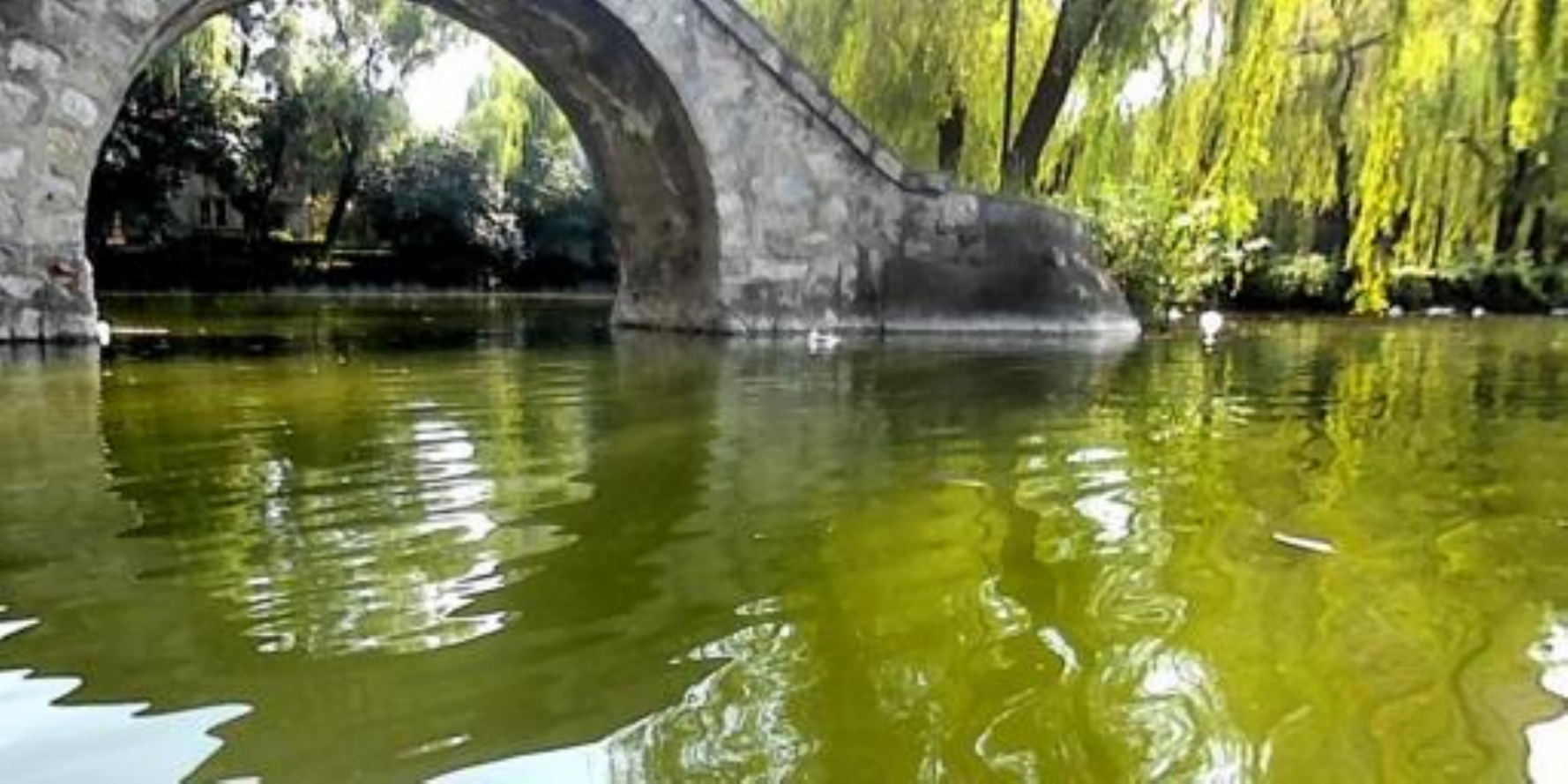} 
\caption{Sun}
\label{fig:Sun}
\end{subfigure}
\begin{subfigure}{0.15\textwidth}
\centering
\includegraphics[width=\textwidth]{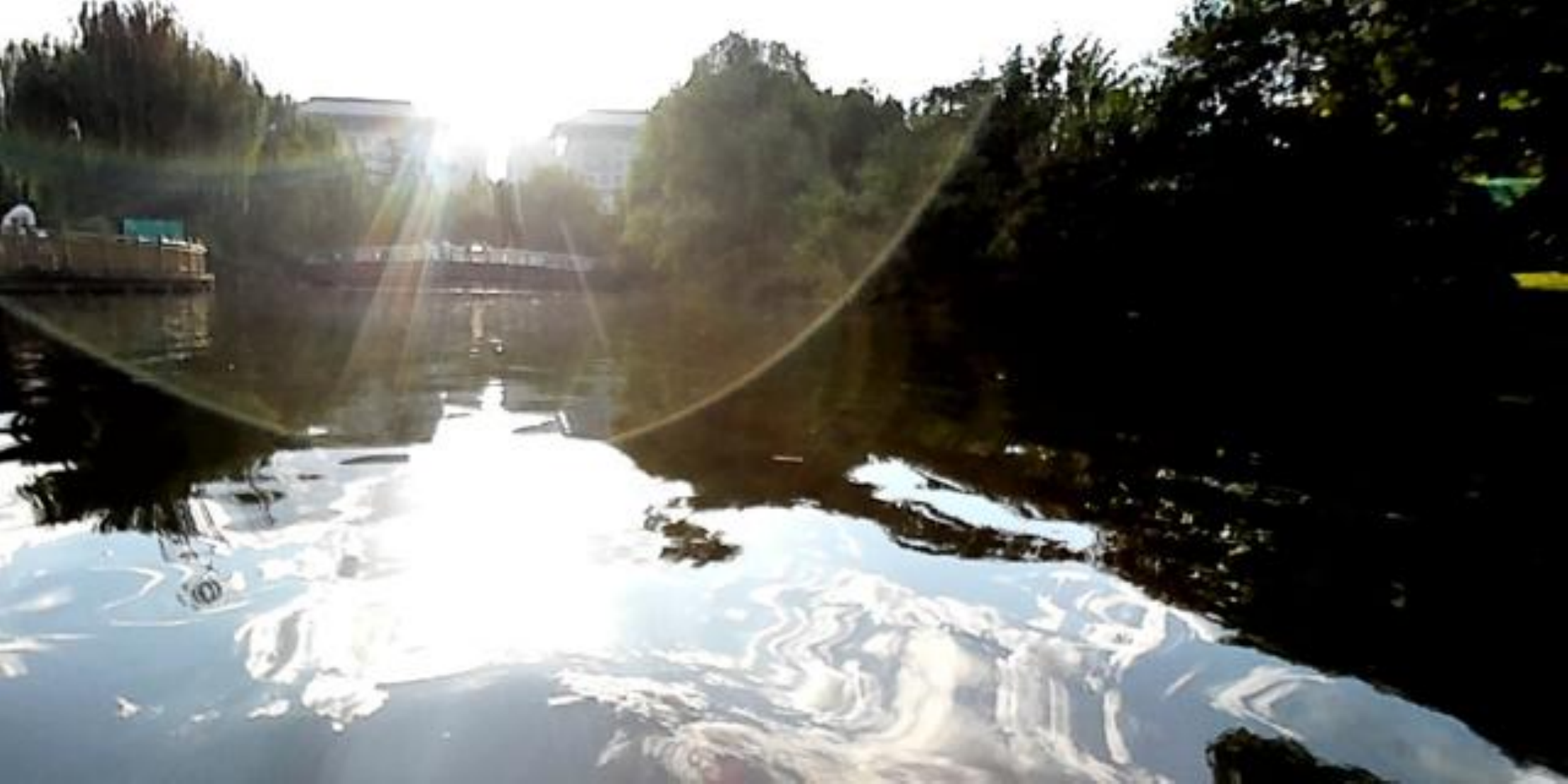}
\caption{Strong Light}
\label{fig:Strong Light}
\end{subfigure}
\begin{subfigure}{0.15\textwidth}
\centering
\includegraphics[width=\textwidth]{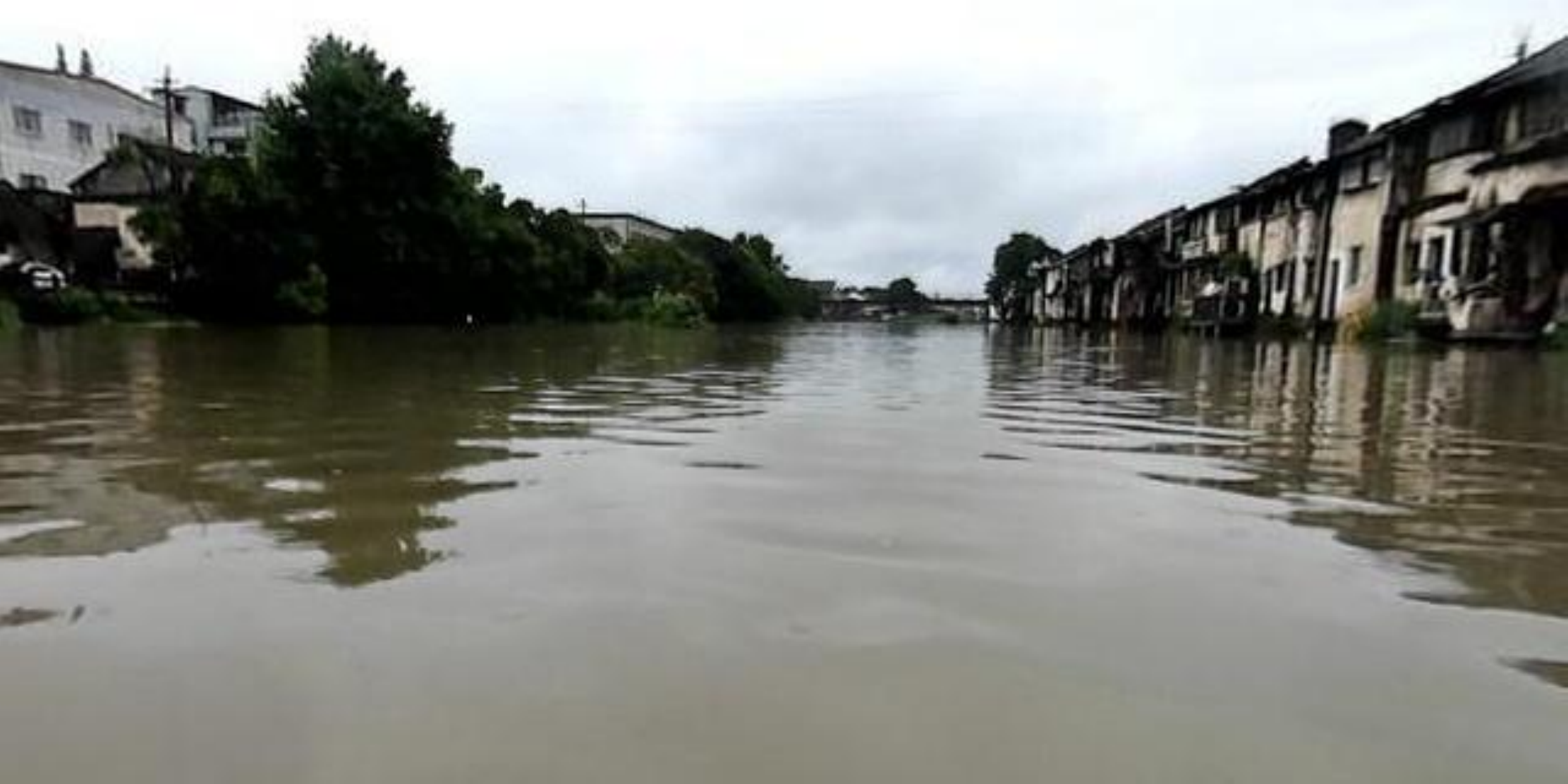}
\caption{Overcast}
\label{fig:Overcast}
\end{subfigure}

\begin{subfigure}{0.15\textwidth}
\centering
\includegraphics[width=\textwidth]{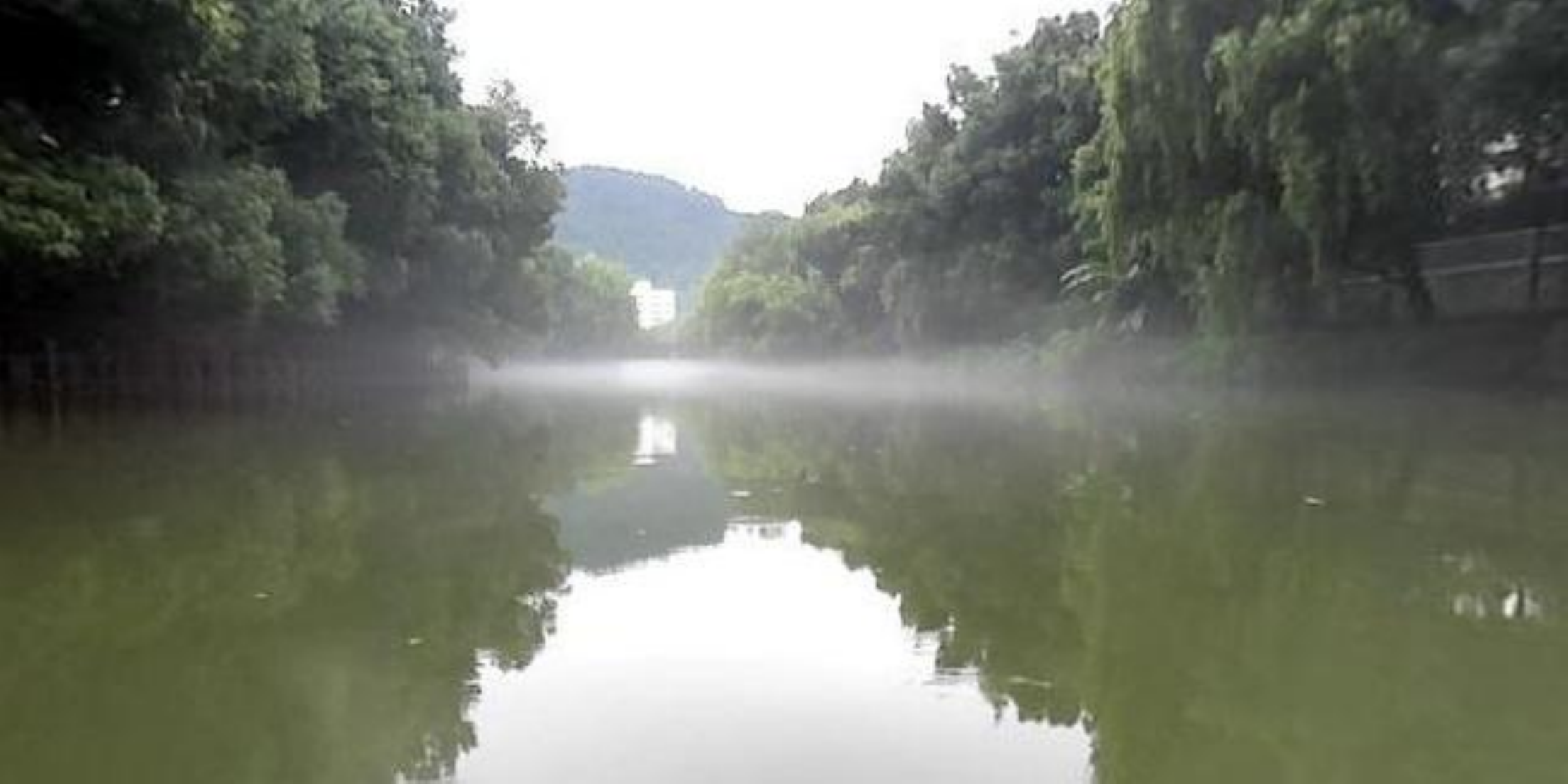}
\caption{Mist}
\label{fig:Mist}
\end{subfigure}
\begin{subfigure}{0.15\textwidth}
\centering
\includegraphics[width=\textwidth]{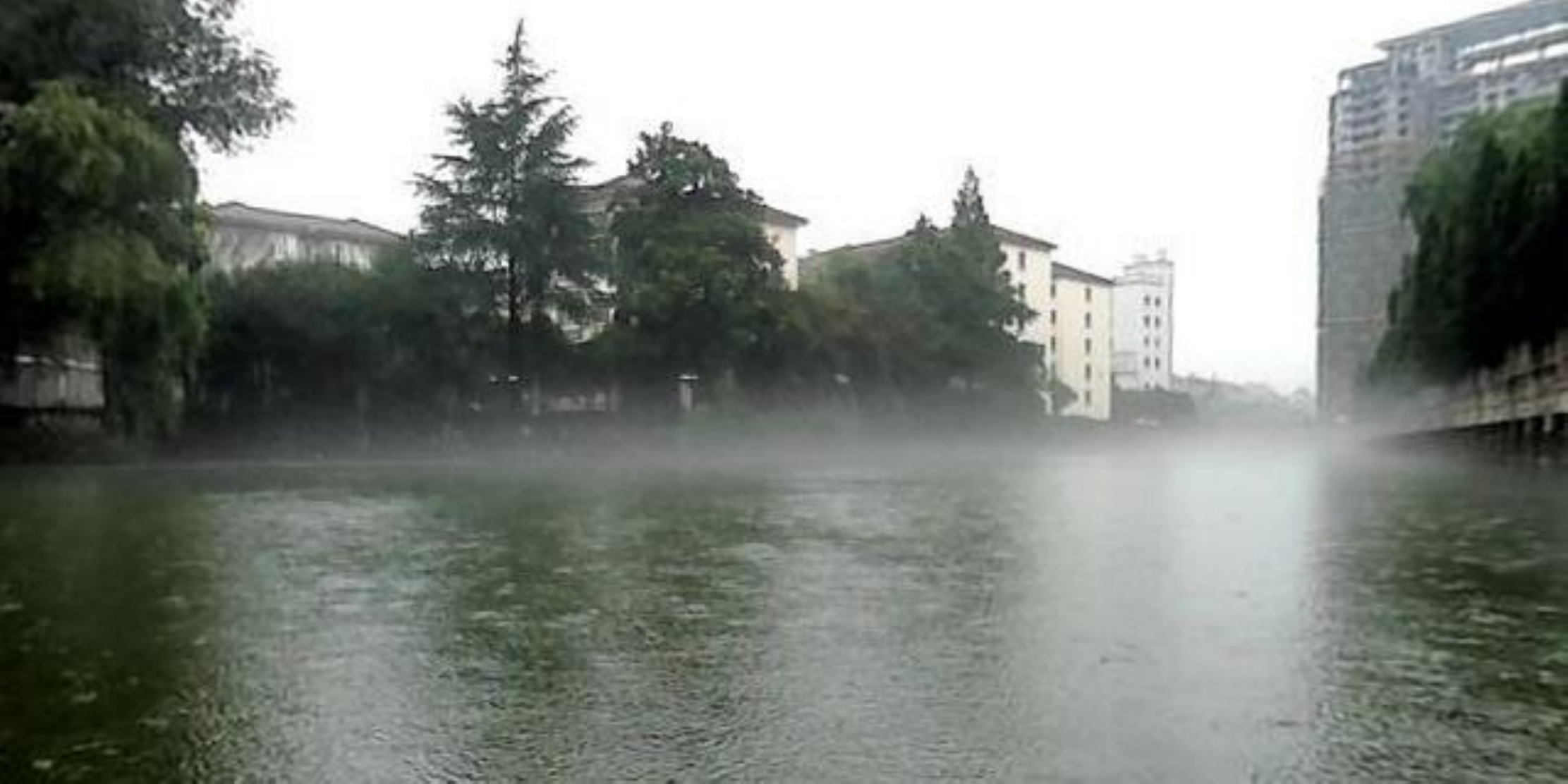}
\caption{Rain}
\label{fig:Rain}
\end{subfigure}
\begin{subfigure}{0.15\textwidth}
\centering
\includegraphics[width=\textwidth]{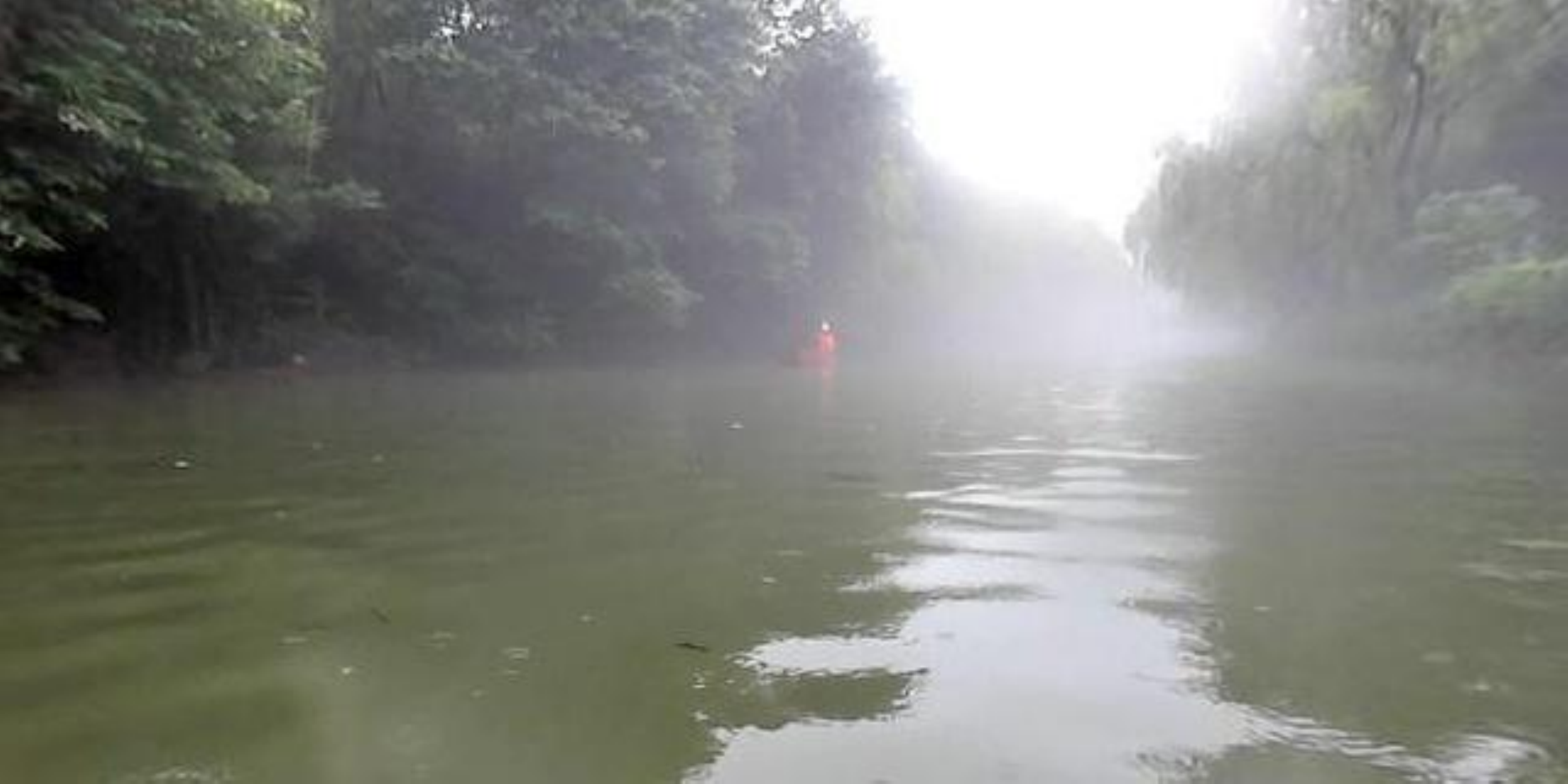}
\caption{Fog}
\label{fig:Fog}
\end{subfigure}

\caption{The dataset is collected under different weather conditions to reveal the challenges of USV perception and navigation in real-world inland waterways.}
\label{fig:weather conditions}
\end{figure}

\begin{figure}
\vspace{0.08in}
\centering
\includegraphics[width=0.48\textwidth]{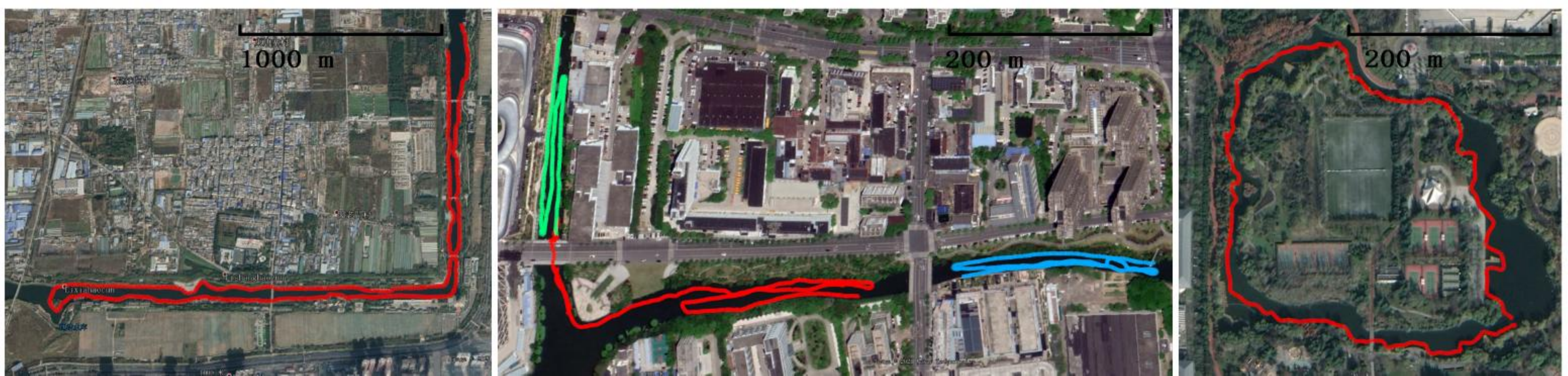}
\caption{Part of the collection trajectories 
% \textcolor{blue}{
(5 of the 27 raw sequences: $X31\_1$ in the image on the left,  $N02\_1$, $N02\_2$ and $N02\_3$ in the middle, $X01\_3$ in the image on the right) are shown here. Our dataset is collected in cities including Xi'an, Ningbo and Hangzhou in China. We name each sequence by the date and location of collection. More details about the data can be found on our website.}
% }
%The collection routes covers more than 26 km in total.
\label{fig:route on map}
\end{figure}

The USVInland dataset collection spans the period from May to August 2020. During this period, our inflatable boat was driven both manually and automatically along inland waterways in typical cities where there are rich hydrographic networks. Considering that harsh weather conditions dramatically increase the difficulties of autonomous driving especially in inland waterways, USVInland contains data collected under various weather conditions as shown in Fig.~\ref{fig:weather conditions} , to cover real-world driving scenes. The waterways we choose are of different widths and lengths with a variety of bankside landscapes. Part of the data collection route is shown in Fig.~\ref{fig:route on map}.

\subsection{Synchronization and Calibration}
For synchronization, at the beginning of data collection, the absolute timestamps of the frames first captured by each sensor are recorded. Then, accurate relative timestamps of each frame from the clock inside each sensor are recorded together with the sensor data. By setting an absolute timestamp as the beginning and referring to the relative timestamps, the data from different sensors can be synchronized.
% \textcolor{blue}{
For autonomous vehicle whose speed can be up to 30-55 km/h, a synchronization accuracy ranging from 0-7 ms can provide a dataset of high quality \cite{sun2020scalability}. As the speed of our boat is much slower (0-6.48 km/s), we believe that the synchronization accuracy of our dataset (a maximum synchronization drift of 25 ms between radar and camera) can meet the requirement of inland waterway scenes.
% }
% For our dataset, the maximum synchronization drift is 50 ms (between Lidar and radar).
% The synchronization drift has a positive correlation with the measurement cycle of each sensor ranging from 10 ms (between IMU and stereo camera) to 50 ms (between Lidar and radar). 

The intrinsic parameters of the stereo camera and the extrinsic parameters of cameras and the IMU inside the same device are from the factory calibration. As in the calibration process used in \cite{meyer2019automotive}, coregistration process of the three sensors (lidar system, camera, radar) is split into two separate processes. First, we use a calibration tool \cite{calibration2019} to calibrate the extrinsic parameters between the camera and lidar system. The approach is based on matching the automatically detected checkerboard corners in the image and the manually extracted lidar point clouds of the checkerboard. To register the radar with lidar, we measure the relative position of the three radars to lidar, which is then optimized by matching the lidar and radar point clouds of several physical targets. The raw dataset we used for calibration is also provided, if some researchers want to obtain calibration results on their own.

%%%%%%%%%%%%%%%%%%%%%%%%%%%%%%%%%%%%%%%% figure_weather %%%%%%%%%%%%%%%%%%%%%%%%%%%%%%%%%%%%%
%%%%%%%%%%%%%%%%%%%%%%%%%%%%%%%%%%%%%%%% figure_weather %%%%%%%%%%%%%%%%%%%%%%%%%%%%%%%%%%%%%
%%%%%%%%%%%%%%%%%%%%%%%%%%%%%%%%%%%%%%%% figure_weather %%%%%%%%%%%%%%%%%%%%%%%%%%%%%%%%%%%%%
%%%%%%%%%%%%%%%%%%%%%%%%%%%%%%%%%%%%%%%% figure_weather %%%%%%%%%%%%%%%%%%%%%%%%%%%%%%%%%%%%%

% For stereo images, we provide both the raw images, undistorted images, and the codes for image distortion correction. 
% We also provide the codes for projecting point clouds of Lidar and Radar into images on our website.
%%%%%%%%%%%%%%%%%%%%%%%%%%%%%%%%%% Figure trace
%%%%%%%%%%%%%%%%%%%%%%%%%%%%%%%%%%Figure trace

%%%%%%%%%%%%%%%%%%%%%%%%%%%%%%%%%%%%%%%% figure_waterline %%%%%%%%%%%%%%%%%%%%%%%%%%%%%%%%%%%%%
%%%%%%%%%%%%%%%%%%%%%%%%%%%%%%%%%%%%%%%% figure_waterline 
\begin{figure}
\vspace{-0.07in}
\centering
\includegraphics[width=0.15\textwidth]{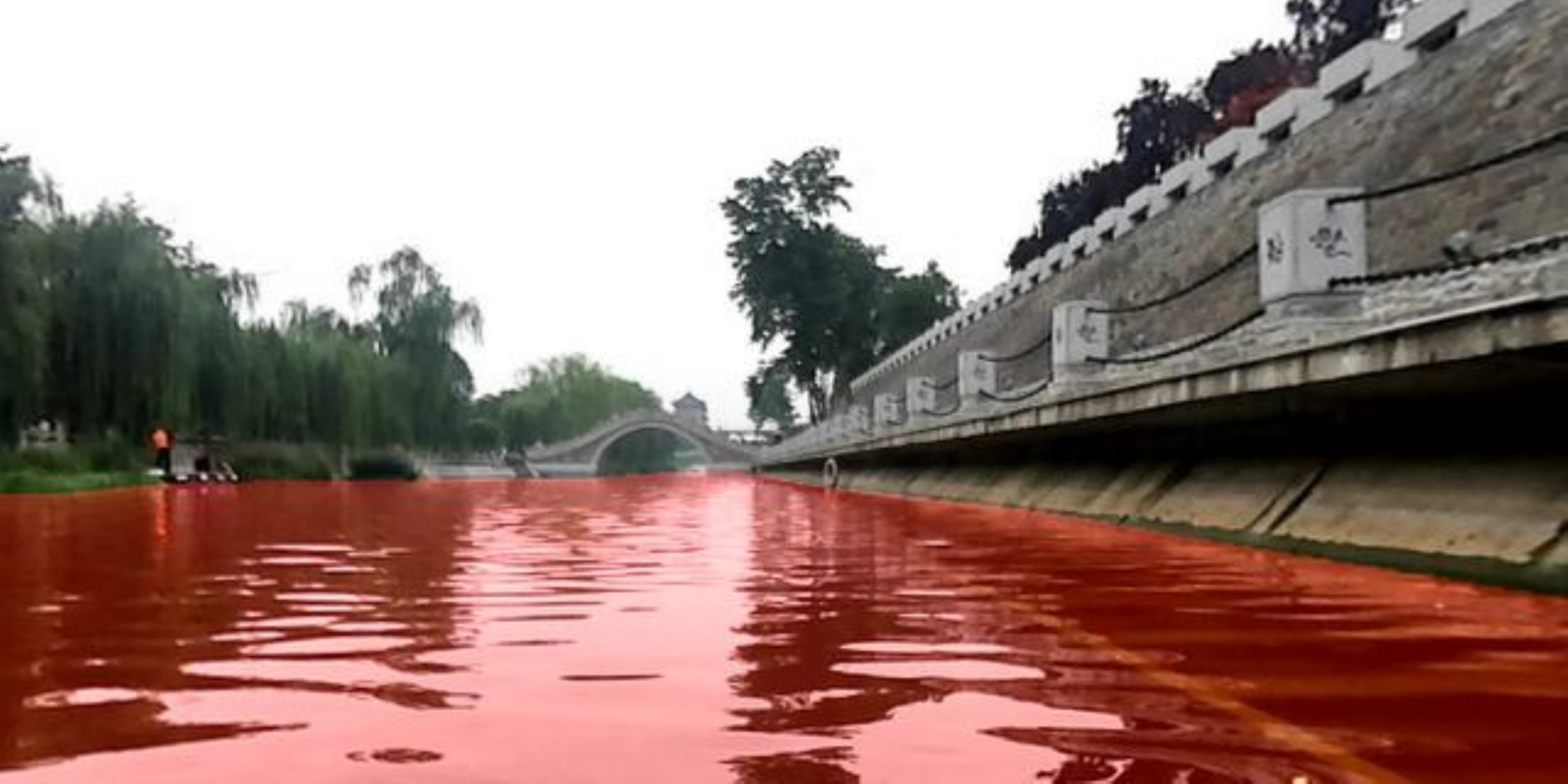}
\hspace*{-0.08in}
\includegraphics[width=0.15\textwidth]{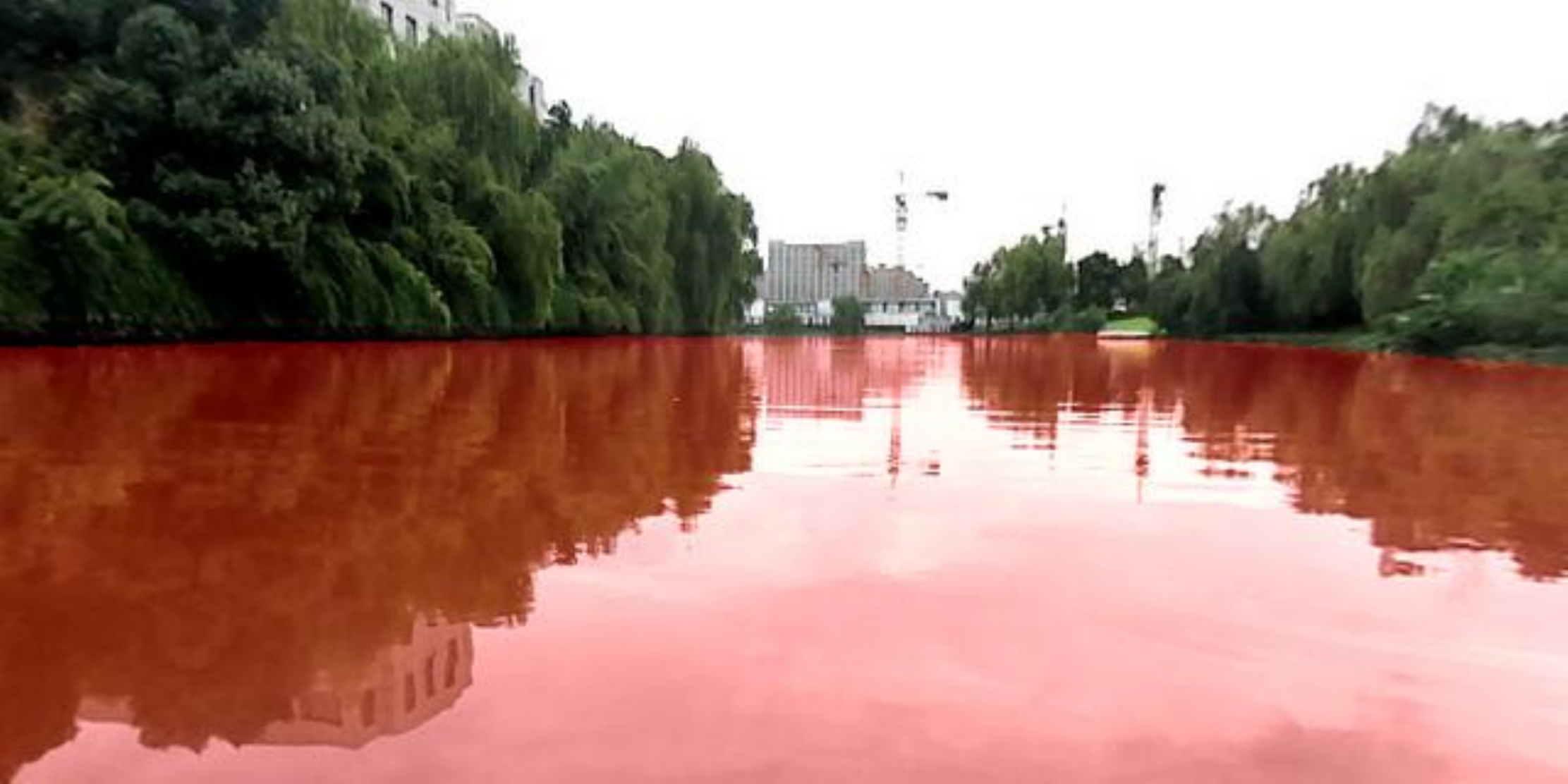}
\hspace*{-0.08in}
\includegraphics[width=0.15\textwidth]{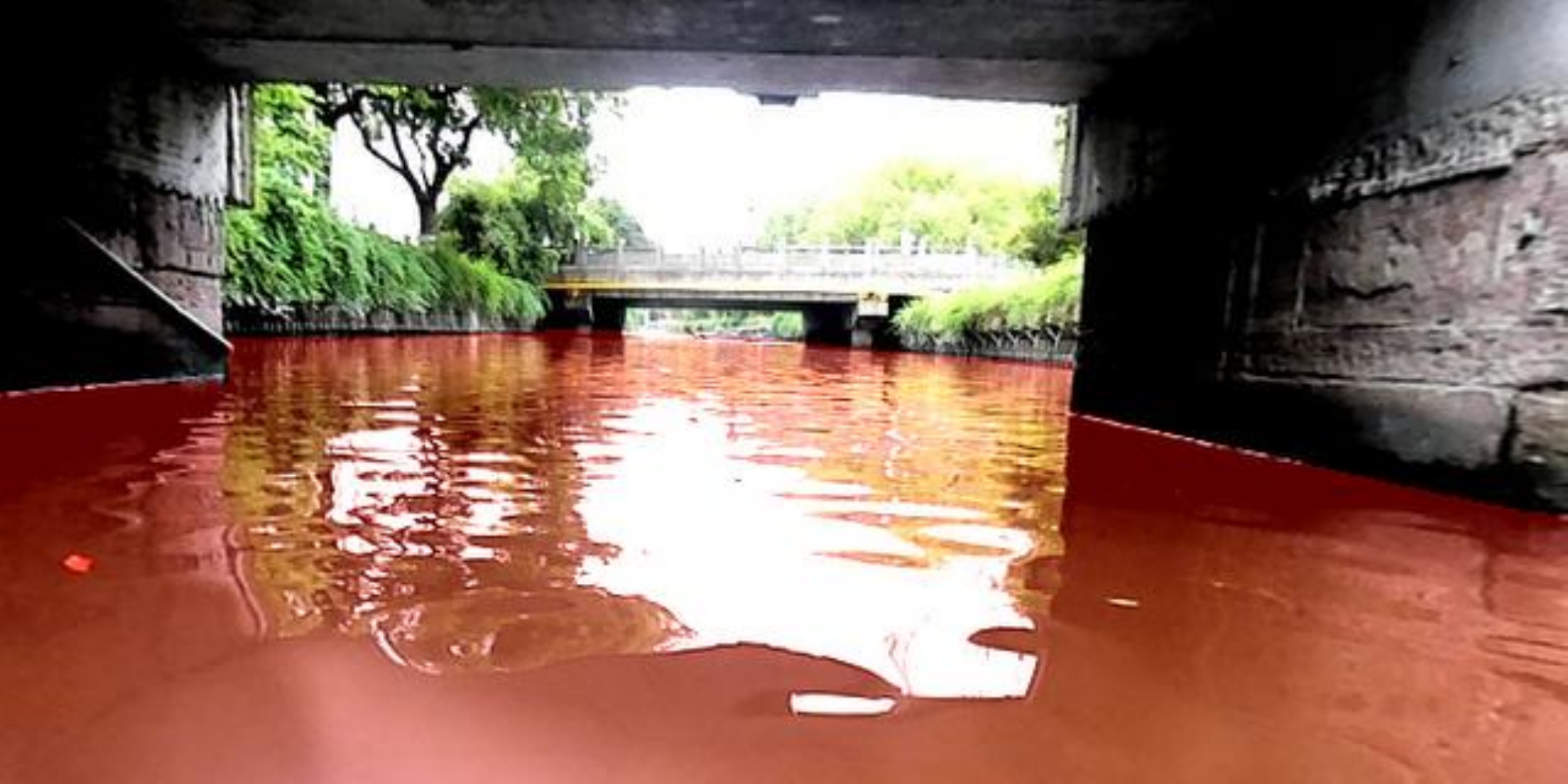}

\vspace*{0.01in}

\includegraphics[width=0.15\textwidth]{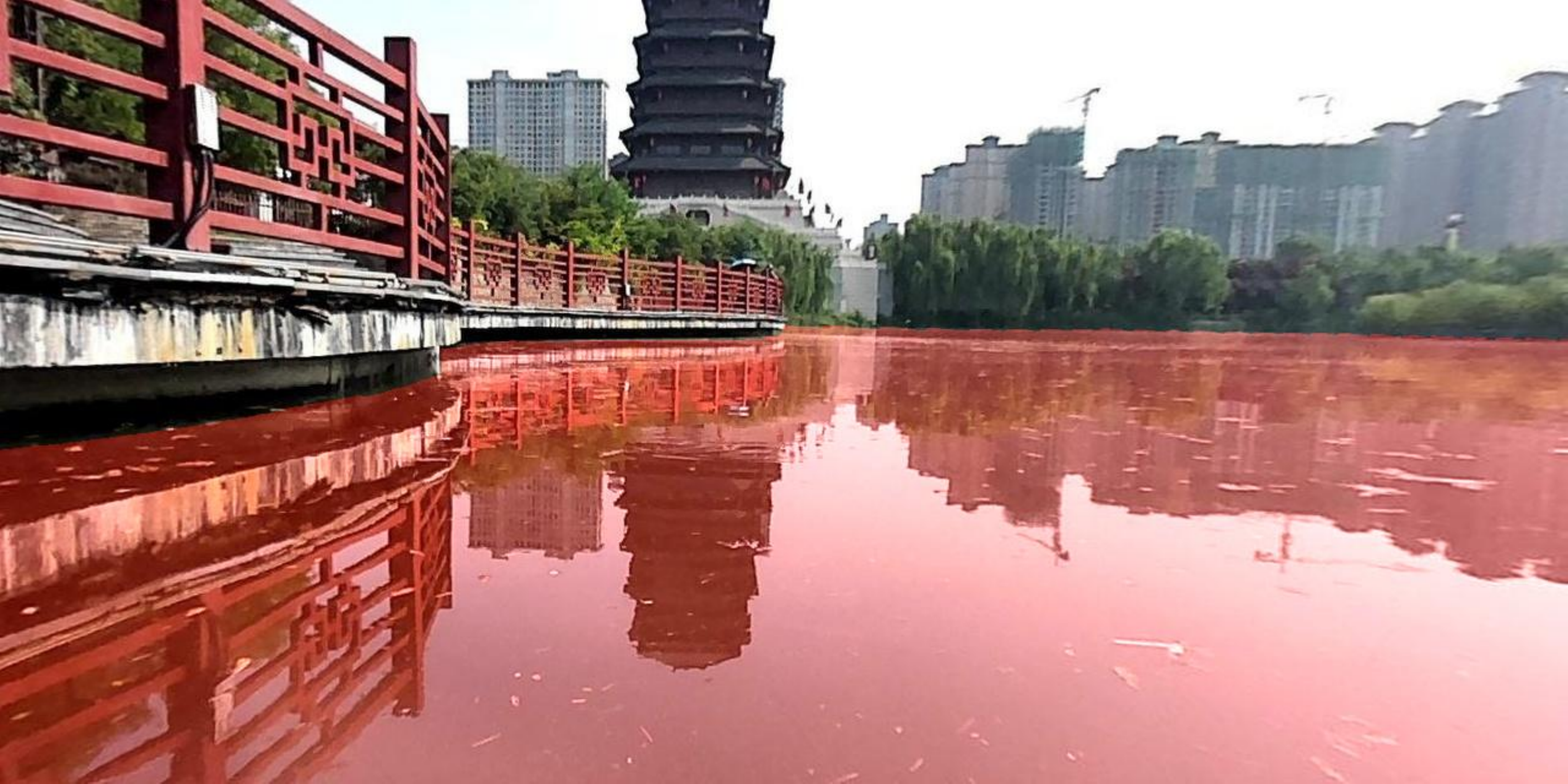}
\hspace*{-0.08in}
\includegraphics[width=0.15\textwidth]{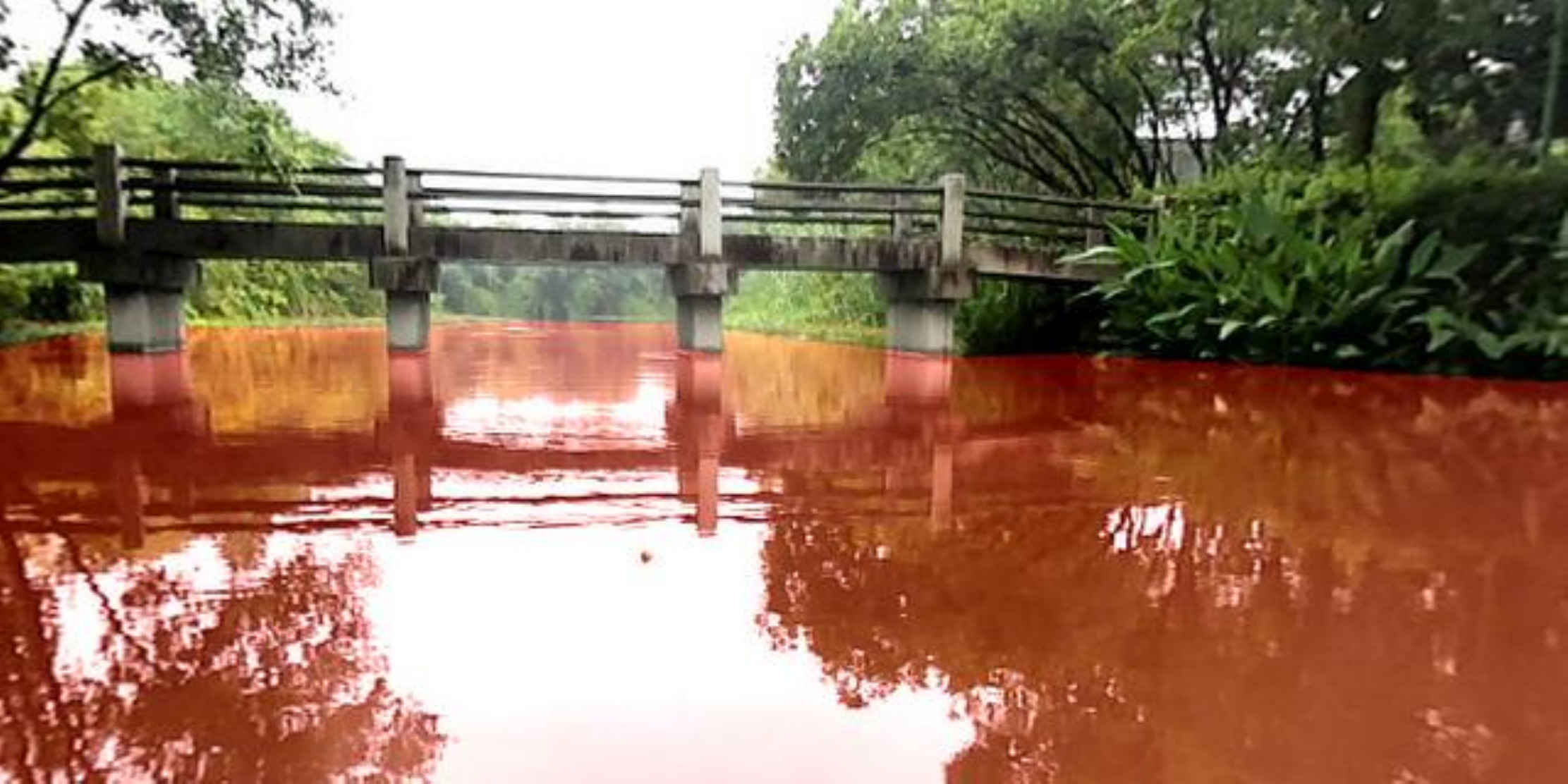}
\hspace*{-0.08in}
\includegraphics[width=0.15\textwidth]{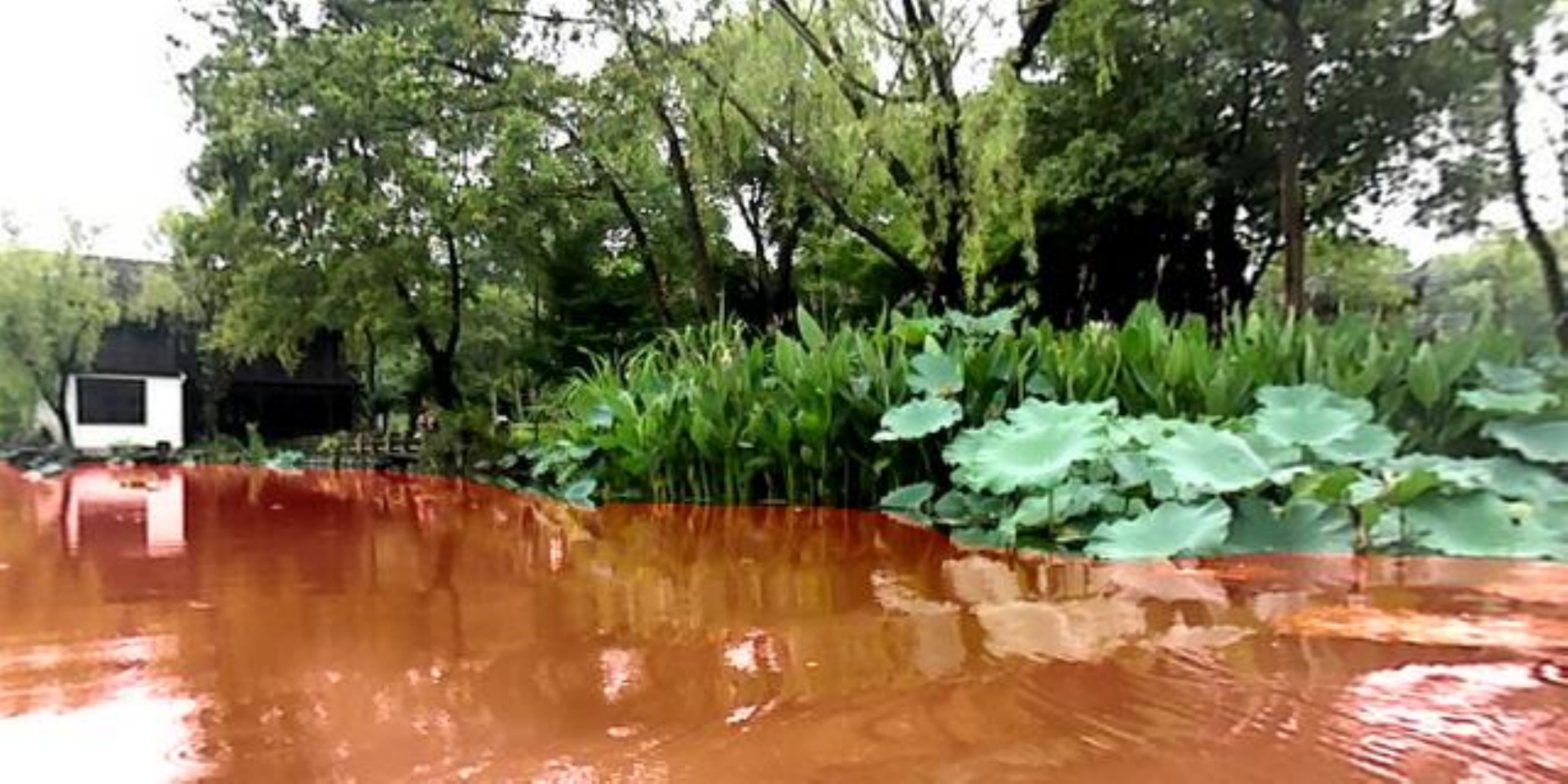}
\caption{The images shown in the figure represent typical waterline structures in our dataset: straight lines, nearly horizontal waterlines in particularly wide waterways, areas under bridges, irregular waterlines, multiple obstacles distributions, and full vegetation on banks. The water area is annotated in red.}
\label{fig:various waterline}
\end{figure}
%%%%%%%%%%%%%%%%%%%%%%%%%%%%%%%%%%%%%%%% figure_waterline %%%%%%%%%%%%%%%%%%%%%%%%%%%%%%%%%%%%%
%%%%%%%%%%%%%%%%%%%%%%%%%%%%%%%%%%%%%%%% figure_waterline %%%%%%%%%%%%%%%%%%%%%%%%%%%%%%%%%%%%%

\subsection{Benchmark Selection}
The raw data we collect cover a trajectory of more than 26 km in total. There are 27 continuous raw sequences collected under different weather conditions. The width of the waterways varies from 10 m to 40 m with a length of the covered waterway ranging from 100 m to 3400 m. For the three tasks, the data are extracted from the raw dataset.

% The statistic of the raw data is show in Fig~\ref{}.

Our \textbf{SLAM} benchmark consists of 33 public sequences in total, which are cut from the 27 raw sequences. The duration of each SLAM sequence ranges from 100 seconds to 2600 seconds. As the GPS signal might be unavailable when the boat is driving in some narrow waterways, the localization precision factor is included in the data during collection to avoid using low-precision location data as a reference. The GPS localization precision factor is recorded as RTK Fixed and RTK Float for different levels of accuracy. Only sequences with GPS data under RTK Fixed mode are selected for the SLAM benchmark.
% The route of all sequences can be described as one-way (23 sequences) %backtracking and round trip (13 sequences). 
% enabling evaluation of algorithms with loop closure detection. 
% Detail information about the path length and collection time of the sequences of the SLAM benchmark shown is Fig.~\ref{fig:SLAM Statistic}.

Our \textbf{stereo matching} benchmark consists of 324 image pairs, which are selected by first downsampling the raw stereo images and applying the similar procedure described in \cite{geiger2012we} based on clustering for the sample diversity. Some extracted obscured and overexposed images are then excluded manually. 

The \textbf{water segmentation} benchmark contains 700 images selected by manually selecting the images of waterways with various waterline shapes. In addition, the images we choose are captured under different weather or lighting conditions. Unlike most datasets used for marine water segmentation in which the waterline is mainly the straight water-sky line, our benchmark covers different landscapes of inland waterways with various waterline structures. Examples of our data are shown in Fig.~\ref{fig:various waterline}. 

\begin{table*}
\vspace{0.08in}
\caption{{Results on some of the sequences in our dataset.}}
% \centering
\begin{center}
\begin{tabular}{c|c|c|c|c|c}
\hline
{Sequence} & {Distance [m]} & {Loop Back} & {Weather} & \multicolumn{2}{c}{{Relative Pose Error*}}\\
\cline{5-6}
{} & {} & {} & {} & {LOAM \cite{zhang2014loam}} & {LeGO-LOAM \cite{shan2018lego}} \\
% {Accuracy\tnote{*}}&{93.120}&{90.714}&{87.435}&{80.858}&{51.197} \\
\hline
{$N02\_6\_1265\_1431$} & {143} & {×} &{Sunny} & {1.8/0.03} & {1.7/0.09} \\
{$H05\_7\_160\_270$} & {135} & {×} &{Rain}  &{2.7/0.03} & {2.8/0.08} \\
{$N03\_3\_605\_760$} & {167} & {×} & {Overcast}  & {1.2/0.04} & {2.5/0.07} \\
{$H05\_8\_30\_205$} & {229} & {\checkmark}&{Rain and Fog} & {6.5/0.17} & {4.8/0.18} \\
{$N03\_2\_80\_536$} & {513} & {\checkmark} &{Overcast}& {2.5/0.09} & {2.6/0.10} \\
{$H05\_9\_115\_700$}&{765} &{\checkmark} &{Mist}  & {2.6/0.02} & {3.5/0.07} \\
% $W06\_1\_290\_845$ & {823} & {\checkmark}&{Rain} & {6.4/0.23} & {6.1/0.19} \\
% $X31\_1\_2040\_4620$ & {2300} &{\checkmark}& {Sunny} & {42} & {2.2} & {2.8} \\
% & {17} & {29}& {38} & {17}  & {20}& {27} & {50}
{$Average$} & {*} & {*} & {*} & {2.9/0.06} & {3.0/0.10} \\
\hline
\end{tabular}
\end{center}
\begin{tablenotes}
\footnotesize
\item[*] %\textcolor{blue}
{*The relative pose errors (RPEs) are measured using segments of trajectories at 5, 10, 50, 100, ..., 400 m lengths: relative translational error in \% / relative rotational error in degrees per 1 m. }
\end{tablenotes}
\label{table:Lidar SLAM}
\end{table*}

%%%%%%%%%%%%%%%%%%%%%%%%%%%%%%%%%% Lidar SLAM
%%%%%%%%%%%%%%%%%%%%%%%%%%%%%%%%%% Lidar SLAM
\begin{figure*}
\centering
\includegraphics[width=0.16\textwidth]{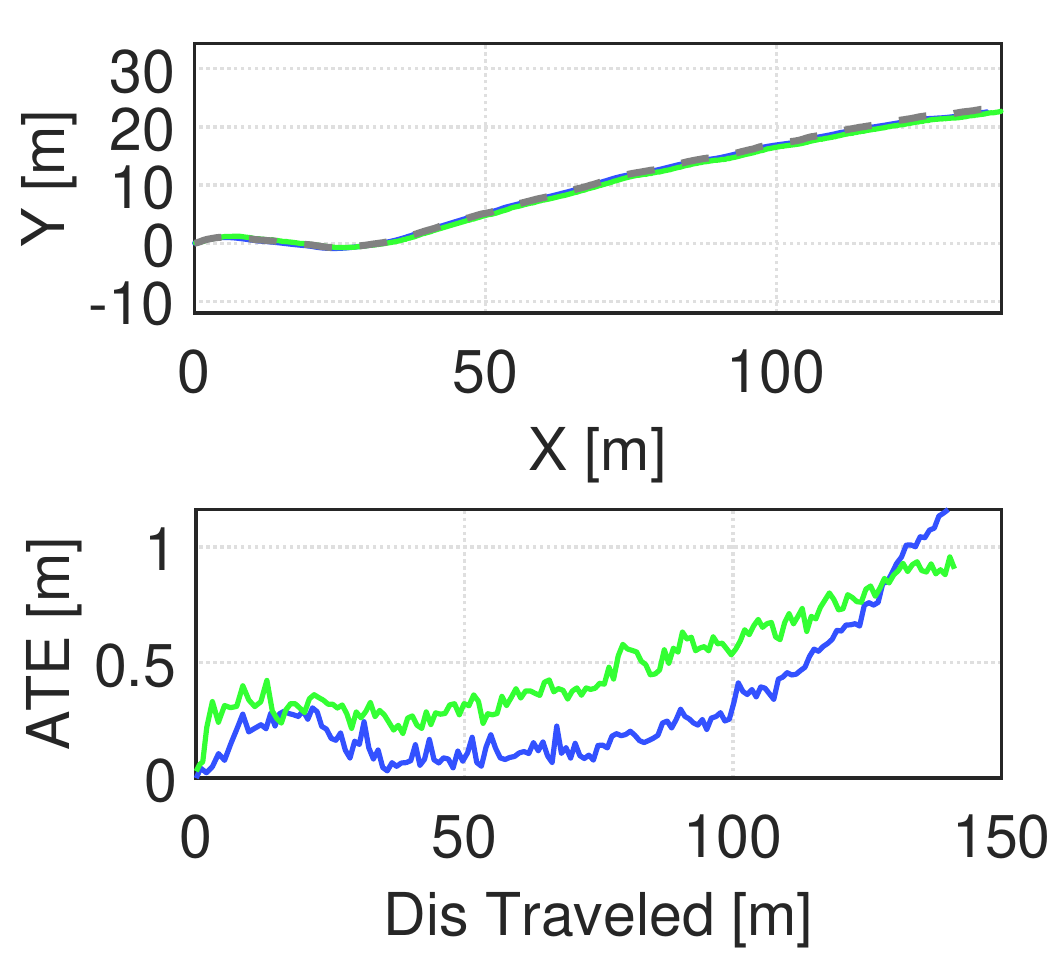}
\hspace*{-0.1in}
\includegraphics[width=0.16\textwidth]{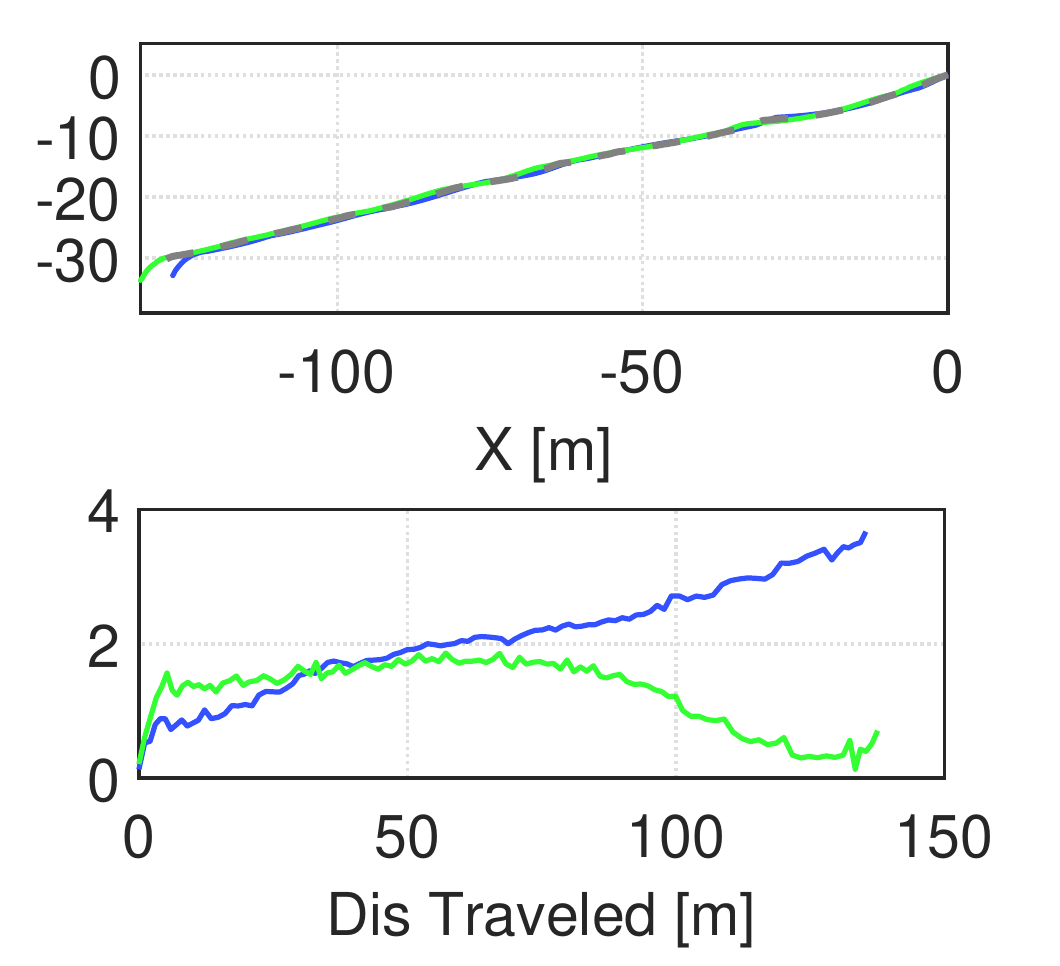}
\hspace*{-0.12in}
\includegraphics[width=0.16\textwidth]{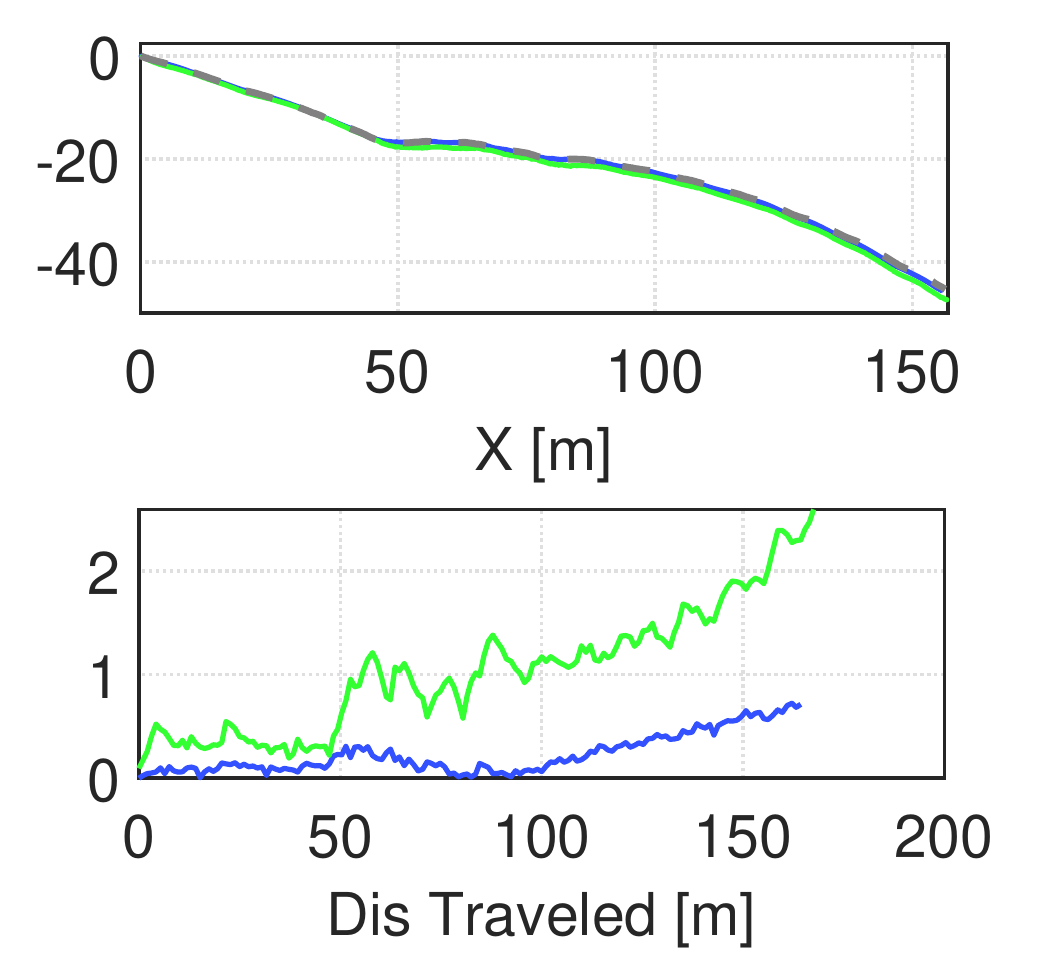}
\hspace*{-0.12in}
\includegraphics[width=0.16\textwidth]{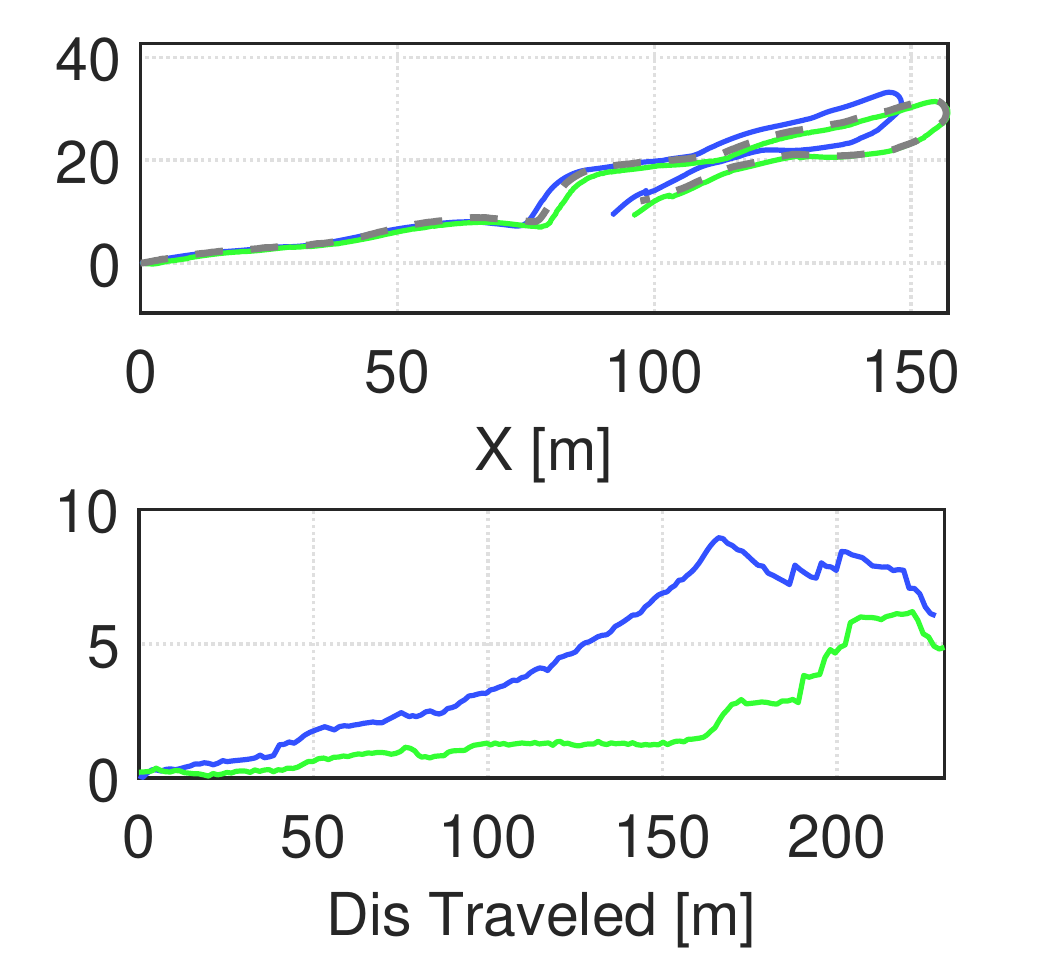}
\hspace*{-0.12in}
\includegraphics[width=0.16\textwidth]{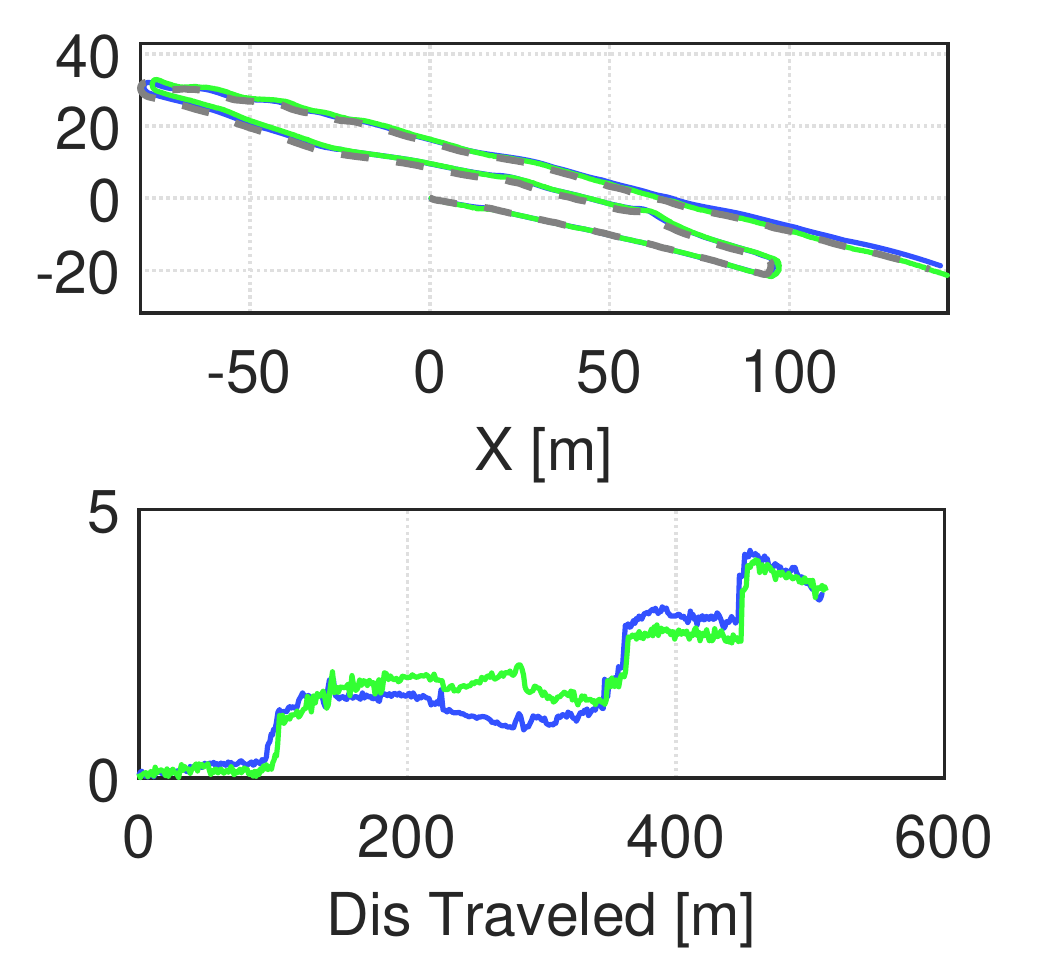}
\hspace*{-0.12in}
\includegraphics[width=0.16\textwidth]{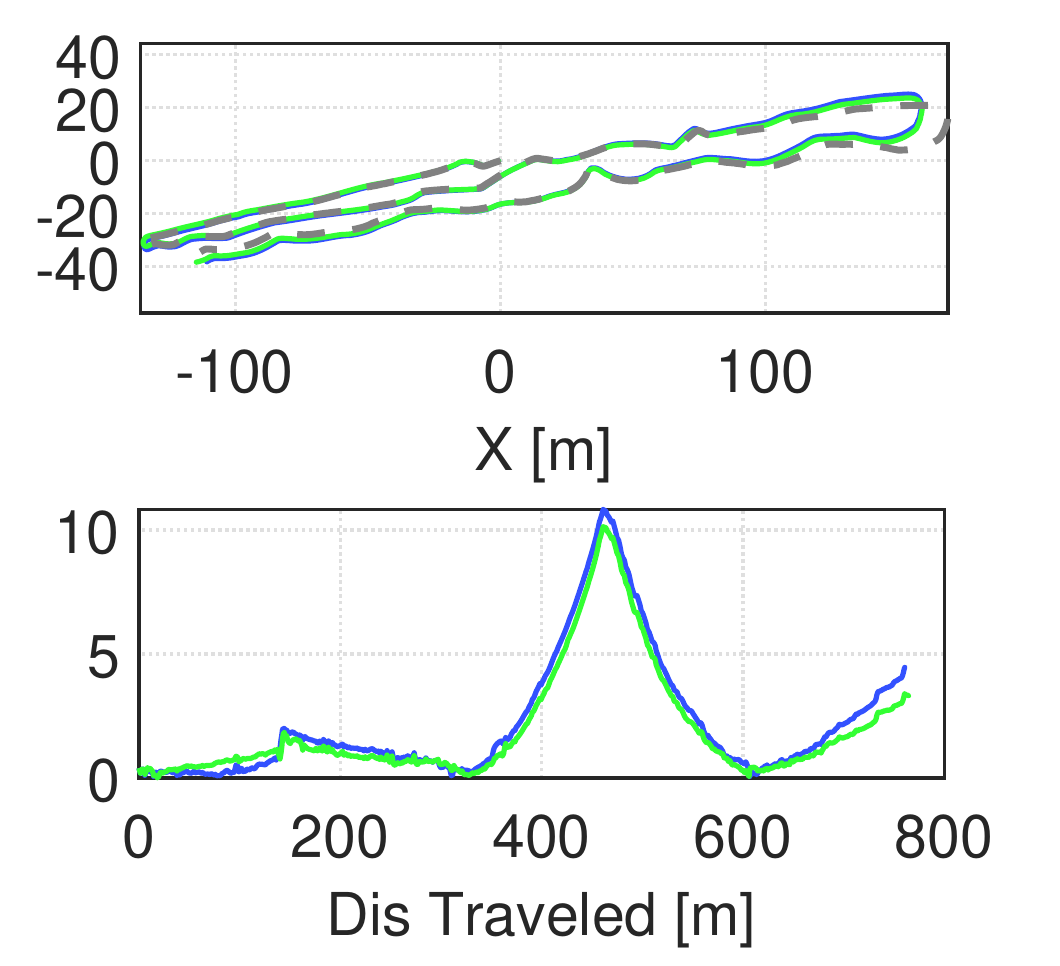}

\caption{ %\textcolor{blue}{
The results of LOAM \cite{zhang2014loam} and LeGO-LOAM \cite{shan2018lego} on 6 sequences (listed in in Table.~\ref{table:Lidar SLAM}) compared with the ground truth trajectories and the absolute trajectory errors (ATEs) as a function of distance traveled. The dashed lines correspond to the ground truth trajectories. The blue and green lines correspond to the results of LOAM and LeGO-LOAM respectively. (The trajectories are rotated to save space.)} %}
\label{fig:Lidar SLAM Experiment}
\end{figure*}
%%%%%%%%%%%%%%%%%%%%%%%%%%%%%%%%%% Lidar SLAM
%%%%%%%%%%%%%%%%%%%%%%%%%%%%%%%%%% Lidar SLAM

\subsection{Ground-Truth Generation}
For \textbf{SLAM}, the ground-truth data come from the output of GPS location and IMU pose information.
%The data from IMU inside of the camera are also provided, including acceleration and angular rate. 
% The orientation groundtruth of visual SLAM can also be generated from the IMU inside of the camera.
% For Lidar and Radar SLAM, the IMU pose can be converted into the coordinate system of the Lidar and Radar. The groundtruth data is then converted into the TUM format \cite{sturm12iros} for easier access and more convenient evaluation.

To generate the ground truth for the \textbf{stereo matching} benchmark, synchronized lidar point cloud data that contain depth information are used. The lidar point clouds in a single frame are relatively sparse when projected onto the image. In this case, we integrate 3D lidar point clouds of adjacent frames based on the iterative closest point (ICP) \cite{chetverikov2002trimmed} and then project the aggregated points onto the image plane for a denser ground truth. No depth ground truth of the water surface area is provided because there is a reduction in the intensity of the reflected light and nearly no lidar point cloud of the water surface is detected.

% \begin{figure}
% \vspace{0.08in}
% \centering
% \includegraphics[width=0.238\textwidth]{picture/SLAM/Lidar/slam_data_1.pdf}
% \includegraphics[width=0.238\textwidth]{picture/SLAM/Lidar/slam_data_2.pdf} 
% \caption{Histogram for trajectory length (left) and lasting time (right) of all sequences for SLAM.}
% \label{fig:SLAM Statistic}
% \end{figure}

To generate ground truth for the \textbf{water segmentation} benchmark, the images are annotated manually by using the \textit{LabelMe} toolbox \cite{labelme2016}. The annotators are asked to label the water area by creating polygons covering the water segment. The annotation quality is ensured as all the annotations are repeatedly verified and corrected.

\section{Experiment}
Based on the three benchmarks, experiments are carried out to show that our dataset is valid and to evaluate the performance of relevant algorithms when applied to inland waterway scenes. 

\subsection{SLAM}
\begin{figure}
% \vspace{0.07in}
\centering
\includegraphics[width=0.46\textwidth]{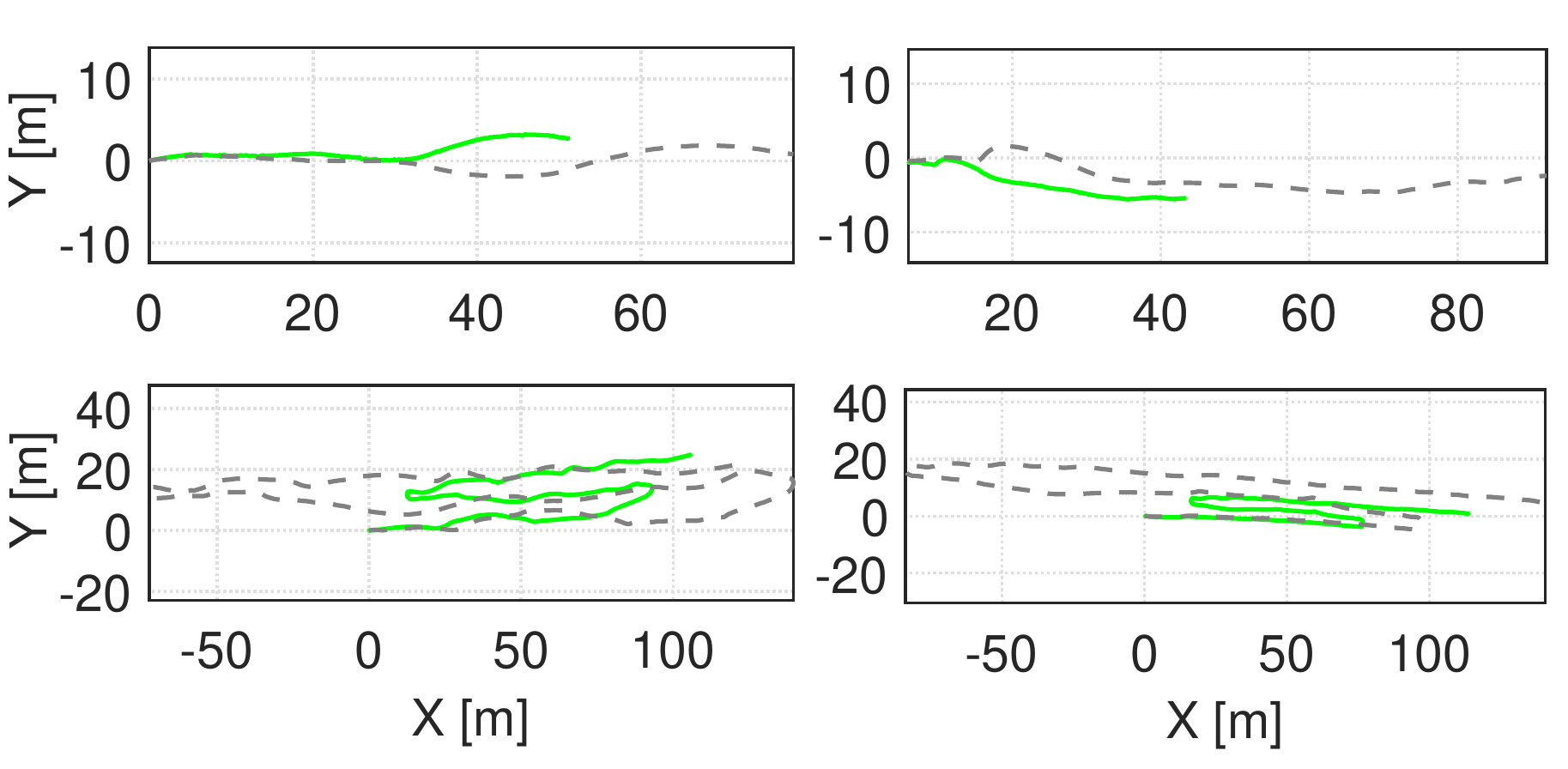}

\caption{%\textcolor{blue}{
Results of ORB-SLAM2 \cite{mur2017orb} on 4 sequences in our dataset: the dashed lines and the green lines correspond to the ground truth trajectories and the results of ORB-SLAM2 respectively. The results differ a lot from the ground truth trajectories. (The trajectories are rotated to save space. The 4 sequences are $W06\_2\_57\_115$, $N03\_4\_440\_523$, $H05\_6\_20\_550$, $N03\_2\_80\_536$ respectively.)}%}
\label{fig:Visual SLAM Res}
\end{figure}

For \textbf{Lidar SLAM}, we evaluated LOAM \cite{zhang2014loam} and LeGO-LOAM \cite{shan2018lego} on some of the sequences in our dataset. %\textcolor{blue}{
We selected 6 sequences collected under different weather conditions. The results are shown in Table.~\ref{table:Lidar SLAM} and Fig.~\ref{fig:Lidar SLAM Experiment}. As the boat sails over the flat and horizontal water surface, the evaluation is done in 2-D space. As can be seen, for sequences collected under harsh environments like rain and fog, there is a decrease in the performance of both LOAM and LeGO-LOAM. %}
The experiment for \textbf{visual SLAM} is based on ORB-SLAM2 \cite{mur2017orb}. %\textcolor{blue}{
We test the algorithms on all the public 33 sequences in our dataset. The performance of the algorithm on our dataset is unacceptable and the estimated results differ a lot from the ground truth. The average relative translation error is more than 20\%. Part of the results and the ground truth trajectories are shown in Fig.~\ref{fig:Visual SLAM Res}.%}

% and the average relative position error ranges from 10.3\% to 23.0\% for ORB-SLAM (without IMU) and ranges from 13.2\% to 21.8\% for VINS (without IMU).
% Oriented FAST and Rotated BRIEF
% As shown in Fig~\ref{fig:Visual SLAM Experiment}, the performance of ORB-SLAM is unstable in real word inland waterway scenes. 
% In inland waterway with distinct bankside features and under a well weather condition, the translation error is relatively smaller. 

%%%%%%%%%%%%%%%%%%%%%%%%%%%%%%%%%% Visual SLAM
%%%%%%%%%%%%%%%%%%%%%%%%%%%%%%%%%% Visual SLAM

\begin{figure}
% \vspace{0.08in}
\centering
% \includegraphics[width=0.23\textwidth]{picture/SLAM/Stereo/fog1.pdf}
% \label{fig:Visual SLAM Fog}
% \hspace*{-0.08in}
% \includegraphics[width=0.23\textwidth]{picture/SLAM/Stereo/Reflection1.pdf}
% \label{fig:Visual SLAM Reflection Features}
\includegraphics[width=0.46\textwidth]{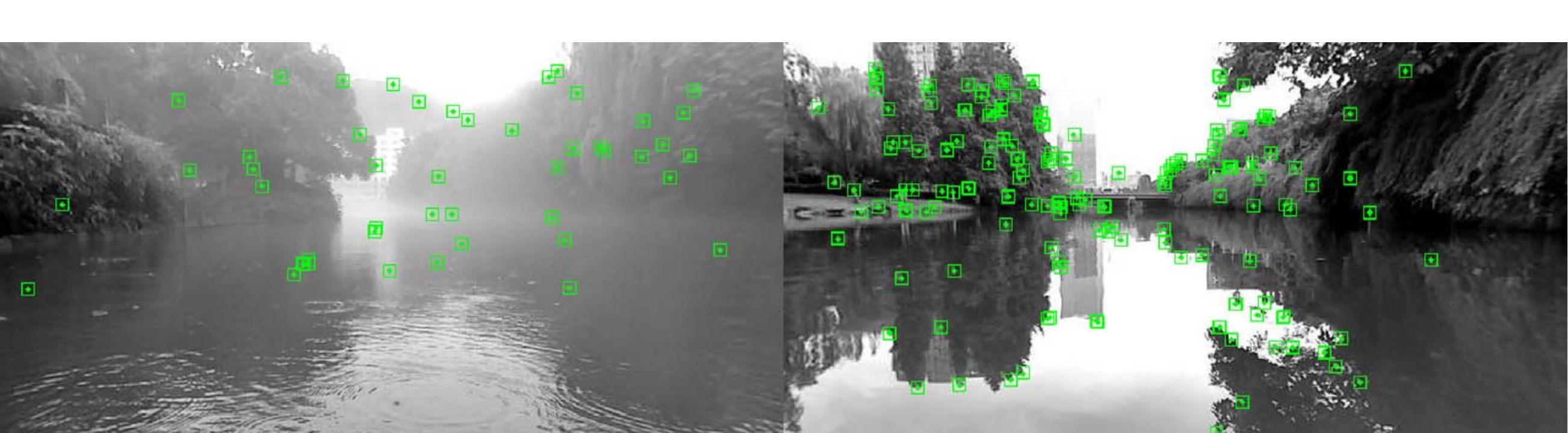}
\label{fig:Visual SLAM Pic}

\caption{Representative scenes for features detected in ORB-SLAM2 \cite{ORB-SLAM2}: the green points show the ORB features. The image on the left shows the detected feature points with fog over the surface. The image on the right shows that the reflection textures are detected as mapping features.}
\label{fig:Visual SLAM scenes}
\end{figure}

%%%%%%%%%%%%%%%%%%%%%%%%%%%%%%%%%% Radar SLAM
\begin{figure}
% \vspace{0.12in}
\centering
\includegraphics[width=0.135\textwidth]{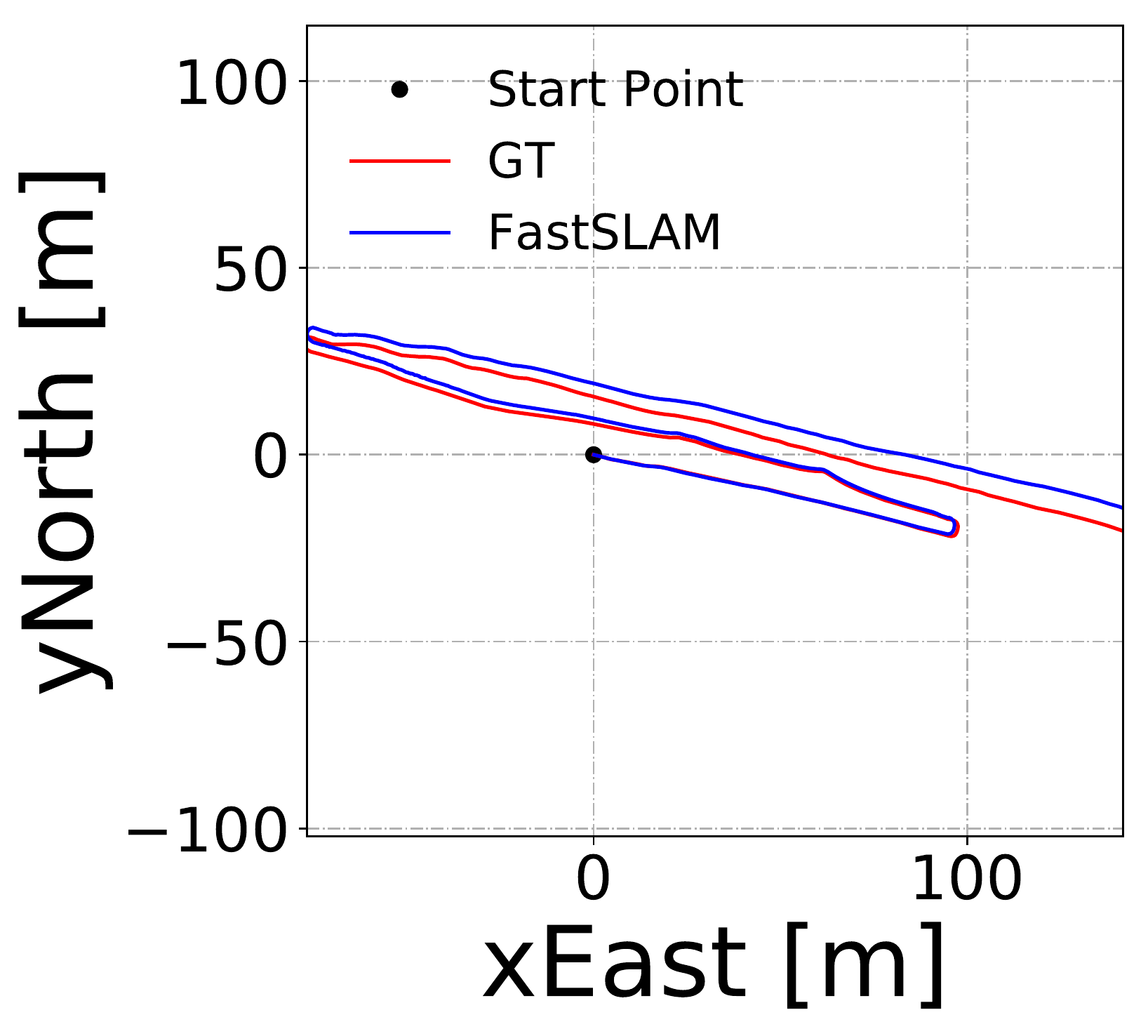}
\hspace*{-0.08in}
\includegraphics[width=0.12\textwidth]{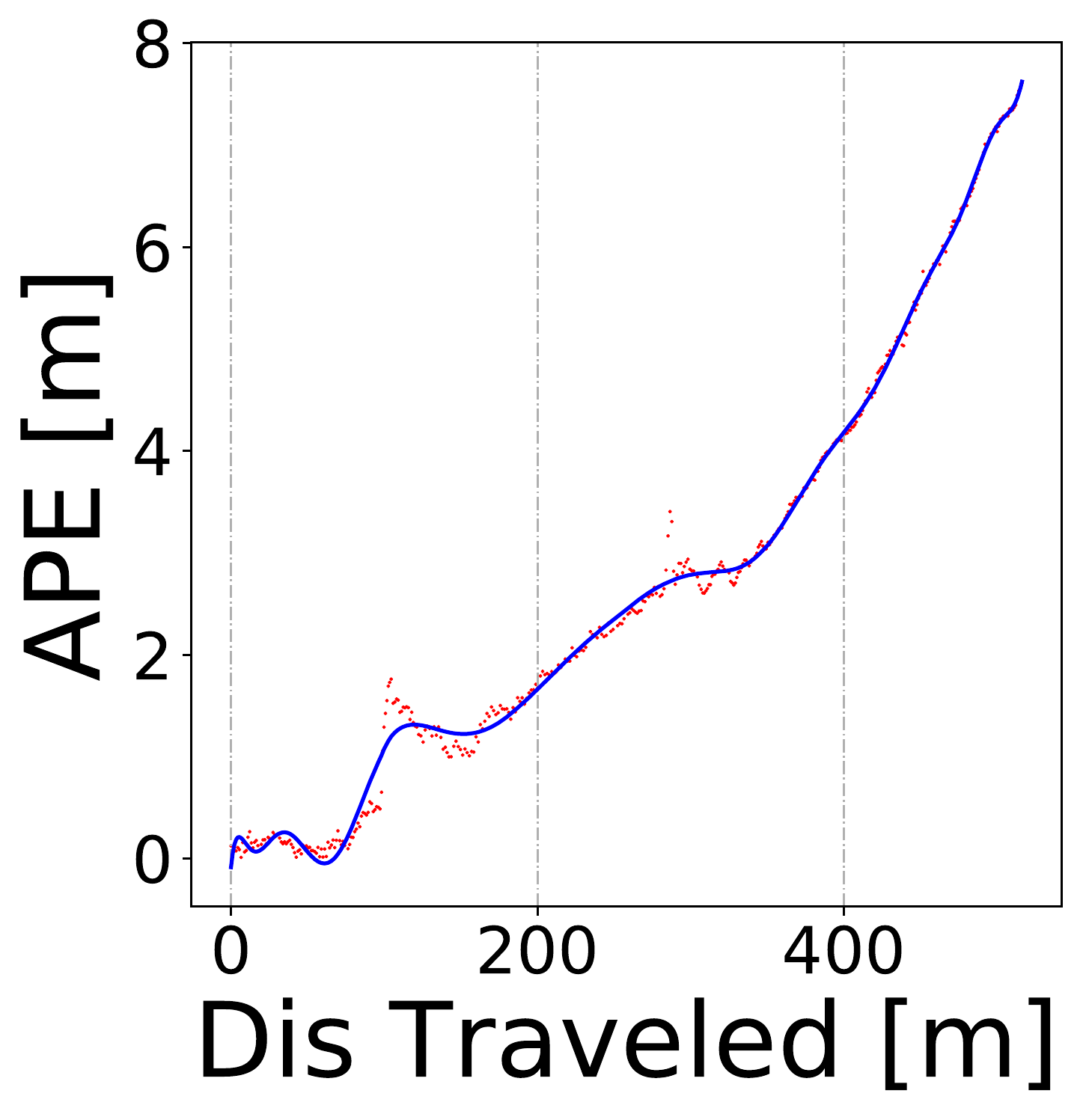}
\hspace*{-0.08in}
\includegraphics[width=0.19\textwidth]{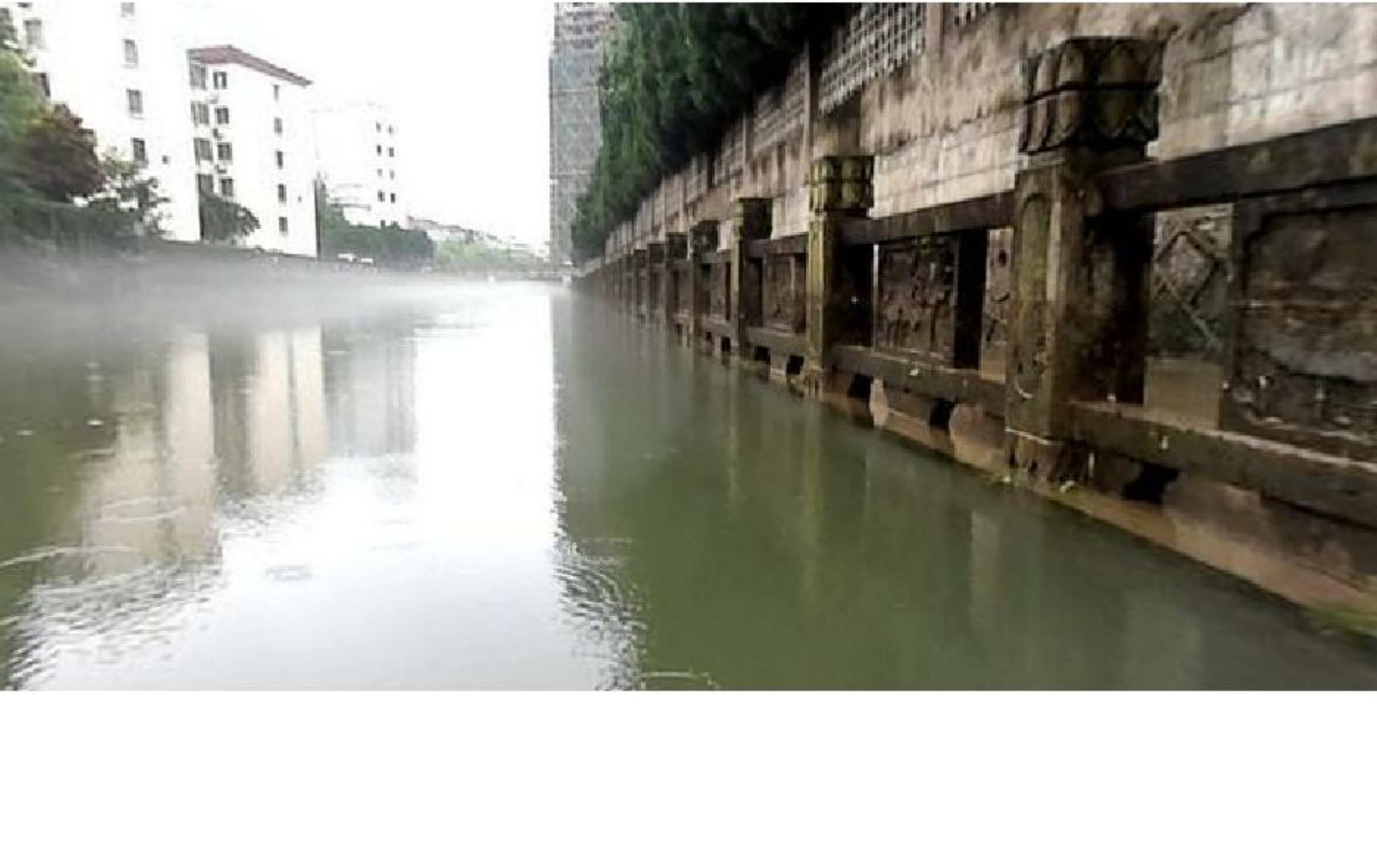}

\vspace*{0.01in}

\includegraphics[width=0.136\textwidth]{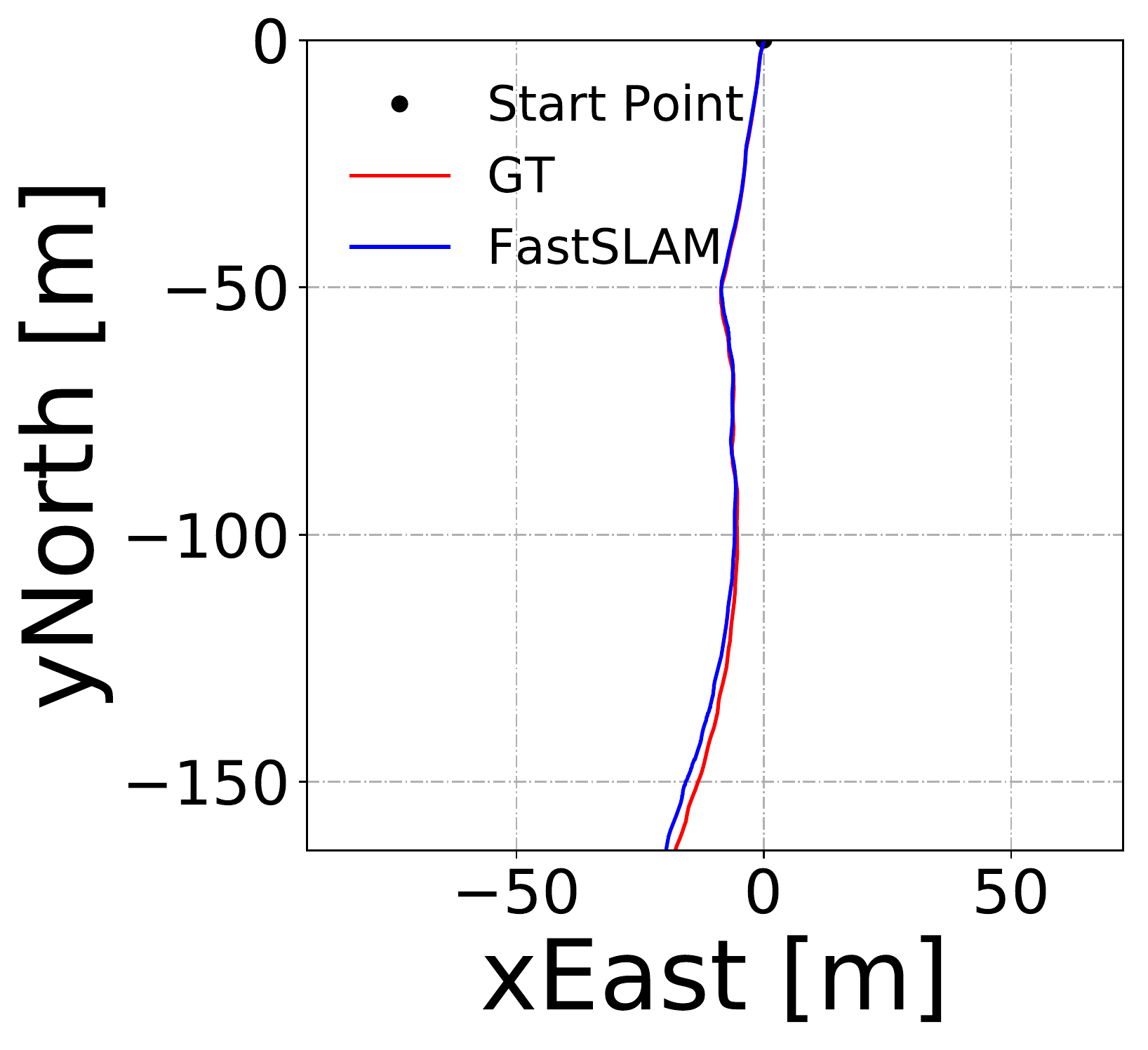}
\hspace*{-0.08in}
\includegraphics[width=0.12\textwidth]{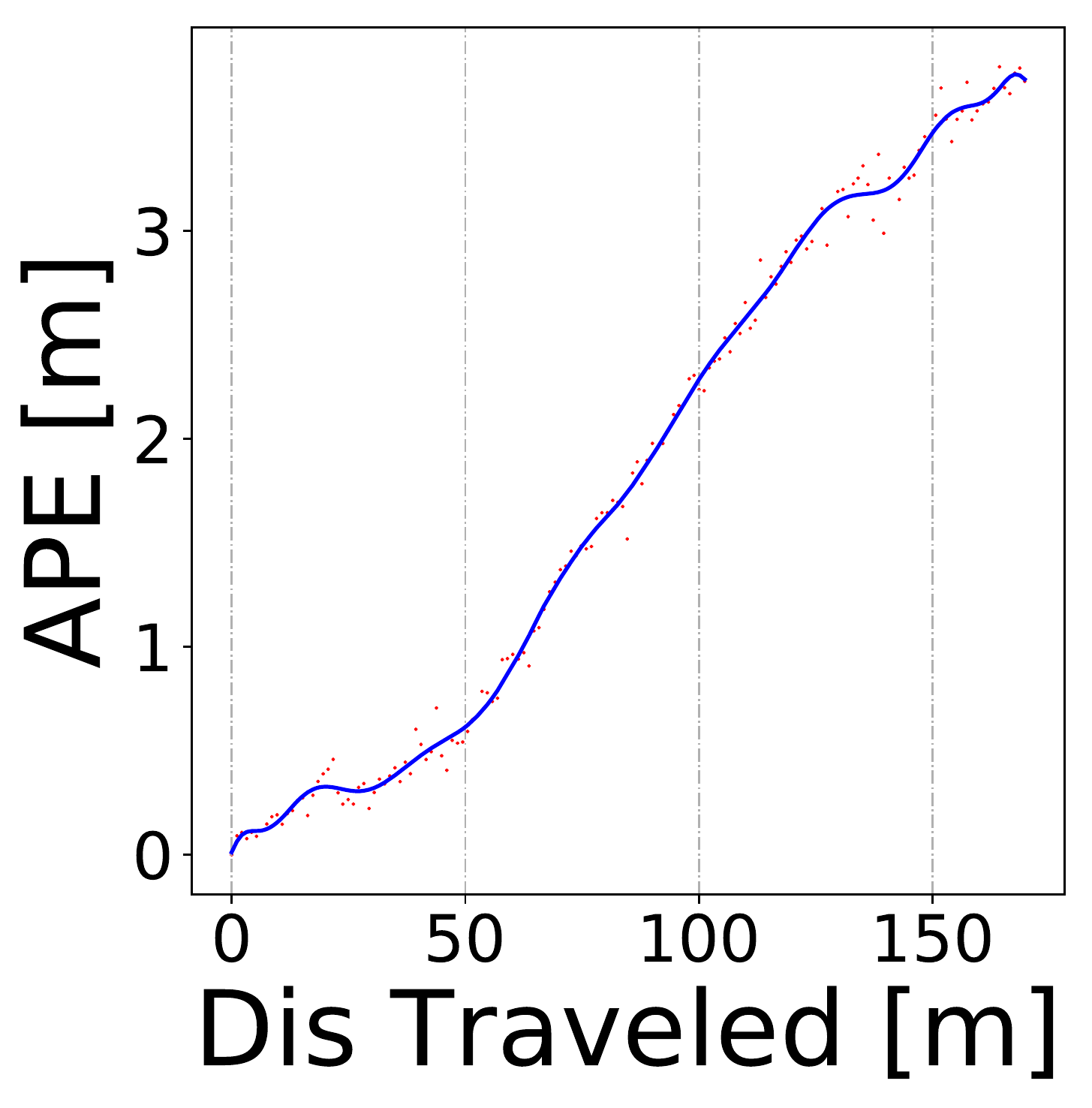}
\hspace*{-0.08in}
\includegraphics[width=0.192\textwidth]{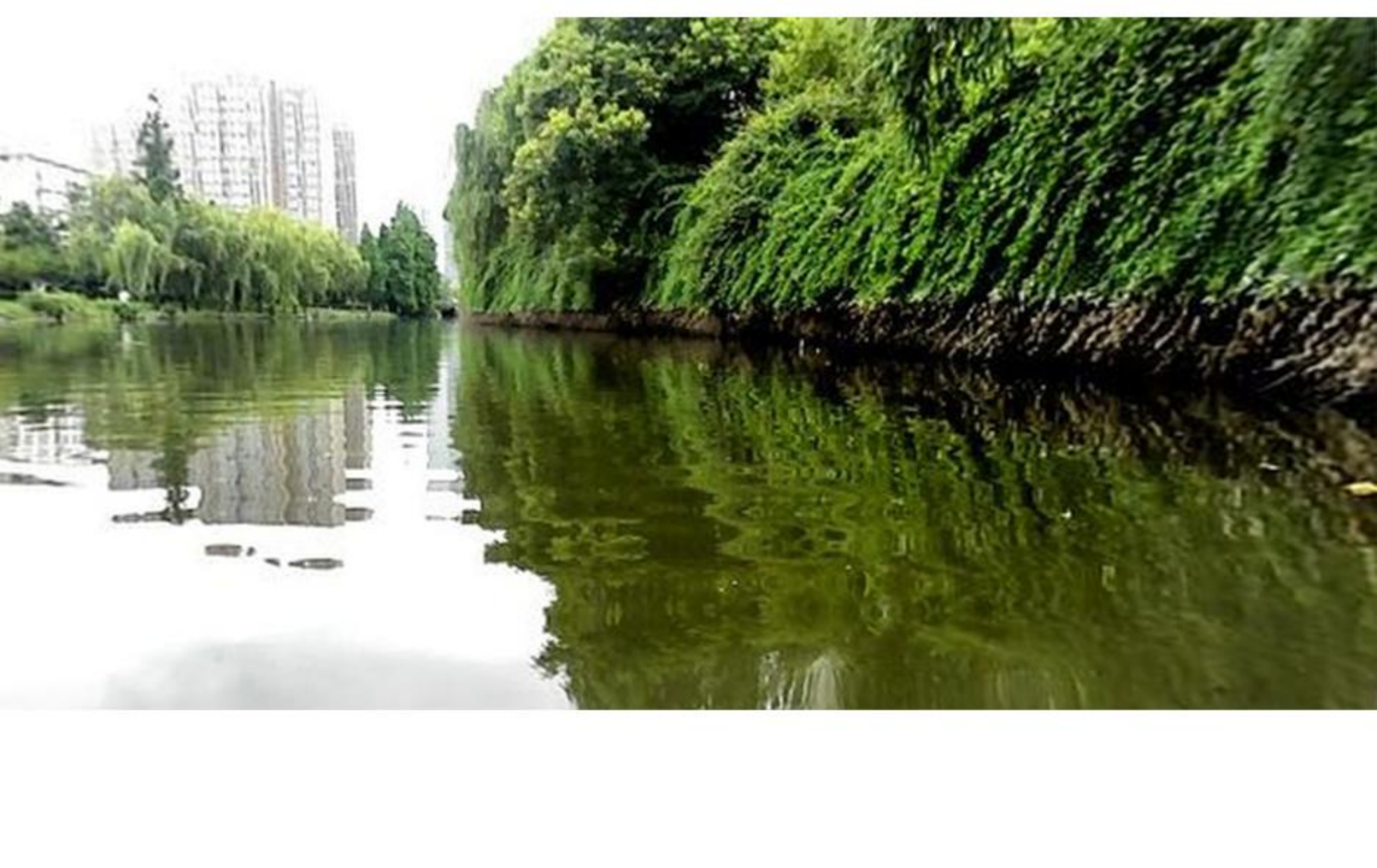}

\caption{
%\textcolor{blue}{
Radar SLAM result: the figures on the left show the estimated results, the ground truth (GT) trajectories, and the absolute trajectory errors (ATEs) as a function of distance traveled. The images on the right show the corresponding scenes. The scene on top contains stone balustrades along the bank.% \textcolor{blue}{
After traveling the same distance, the ATE of the sequence on the top is much smaller than ATE of the one at the bottom. (The two sequences are $H05\_9\_115\_700$ and $H03\_3\_605\_760$ respectively. The red dots are ATEs and the curves are polynomial fitting results of the dots.)}%}}
\label{fig:Radar SLAM Experiment}
\end{figure}

For visual SLAM, it can be observed from Fig.~\ref{fig:Visual SLAM scenes} that, on the one hand, the fog over the water surface can reduce visibility and lead to difficulties in detecting feature points. % \textcolor{blue}{
On the other hand, as mentioned in \cite{chambers2011perception}, the lack of salient features in short range (mainly extracted from river bank) reduces the performance of feature based algorithms. In our experiment, in inland waterways that are wide and when the vehicle is driving in the middle of the waterway, most of the extracted features are from the reflection on the water surface and are far away. The errors in depth estimation for the feature points at relatively far distances can lead to errors in relative pose estimation between adjacent frames. In addition, in some cases, the water wave can distort the water reflection, which makes the keypoints in the reflection unstable in adjacent frames. In this case, the estimated result differs greatly from the ground truth. %}
% both in scale and in shape as shown in Fig.~\ref{fig:Visual SLAM Experiment}\subref{fig:scale2}.
% and the reflections can not be treated as the keypoints because the over water surface the conventional transformations among the view angles of two cameras, the disparity map, and the depth of the feature points are infeasible.
% By aligning the scale, the estimated results better match the groundtruth trajectories as shown in the Fig.~\ref{fig:Visual SLAM Experiment}\subref{fig:scale1}. 
% , the result of ORB-SLAM differs a lot from the groundtruth in scale.
%We expect these challenges for visual-based SLAM on inland water surface could be solved.

For \textbf{radar SLAM}, we evaluate the algorithm based on FastSLAM proposed by Schuster \textit{et al.} \cite{schuster2016robust} on our dataset. The radar data are the fusion of the point clouds from 3 millimeter-wave radars, which are mounted at the front and two sides of the platform for more comprehensive information around the vehicle. %\textcolor{blue}{
As the detection range of radar is limited, we select 8 sequences in which the boat sails close to the bankside or the rivers are narrow (the 8 sequences we selected are marked in our website). Testing on these sequences, we obtain a relative translational error of 3.95\% (averaged over trajectories of 5 to 200 m length).%}

The results on two of the selected sequences are presented in Fig.~\ref{fig:Radar SLAM Experiment}. It can be observed that the algorithm achieves a better performance when applied to scenes with distinct targets on the waterway sides for radar, such as the stone balustrade, which contains metals.
% \textcolor{blue}{
However, as shown in the first sequence of Fig.~\ref{fig:Radar SLAM Experiment}, when the boat turned around (at the turning point in the first sequence) and sailed in the middle of the waterway, being further to the waterway sides, the drift between the estimated result and ground truth grew a lot as the number of valid targets decreased.

\subsection{Stereo Matching}
\begin{figure}
\centering
\includegraphics[width=0.24\textwidth]{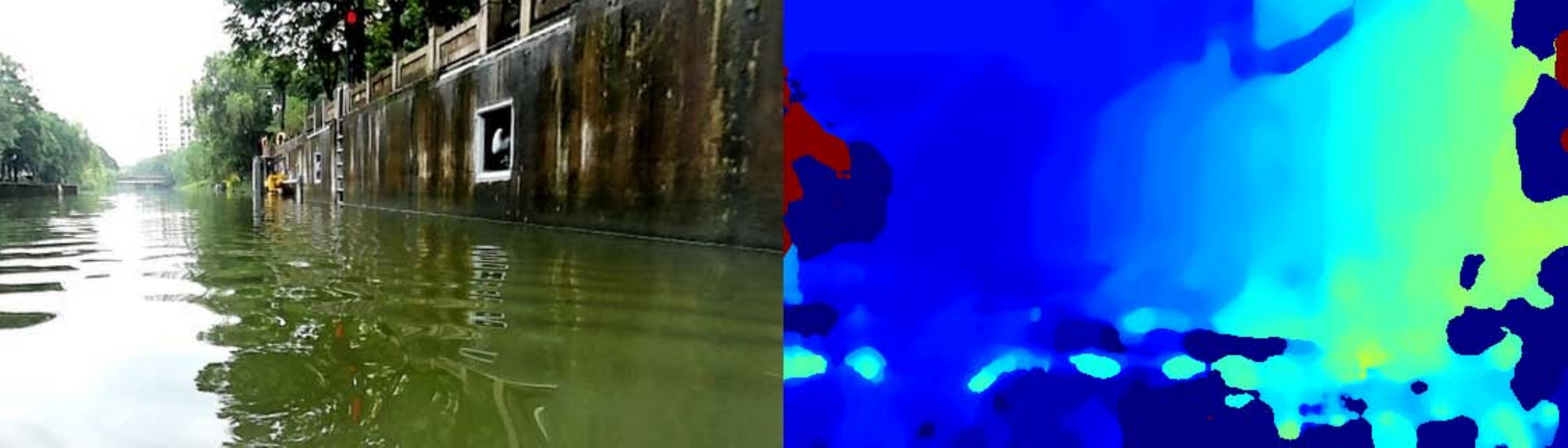}
\hspace{-0.08in}
\includegraphics[width=0.24\textwidth]{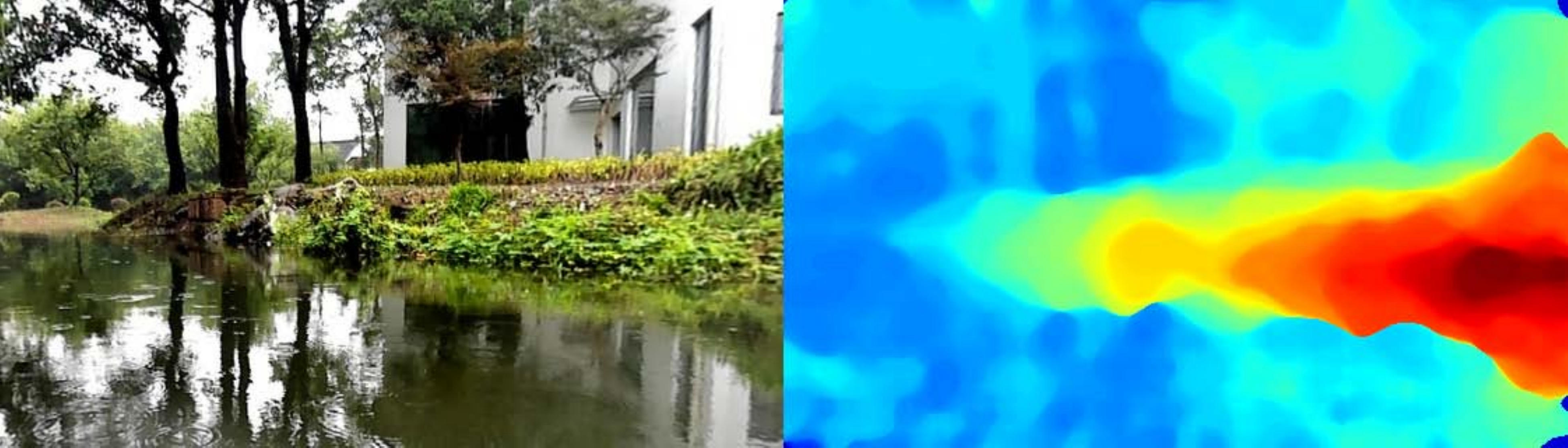}
\caption{
% Histogram of average disparity error of all 180 image pairs (top).
Disparity map of relatively unacceptable (left) and good (right) results for stereo matching based on SGM.}
\label{fig:Stereo Matching}
\end{figure}

For stereo matching, we apply SGM \cite{hirschmuller2007stereo} on the 180 low-resolution (640×320) image pairs in our stereo matching benchmark. The SGM result is compared with the ground-truth disparity map generated by transforming the depth information from the lidar point clouds. Only pixels with ground-truth data are included in the evaluation. We calculate the error percentage as a function of the disparity error threshold. The result is shown in Table.~\ref{table:Stereo Matching SGM}. The SGM performance is not good when applied to real-world scenes of inland waterways. Examples of the images and corresponding disparity maps generated by SGM are shown in Fig.~\ref{fig:Stereo Matching}. It should be emphasized that the disparity map of the water surface areas cannot be accurately estimated by common algorithms because the disparity map of water surface areas is estimated based on the keypoints extracted from the reflection of the bankside objects. At present, we evaluate the stereo matching algorithm mainly 
on the non-water areas, as no ground-truth data for the water area are provided and we think that to enable collision avoidance, the depth information of the bankside area is more important.

\subsection{Water Segmentation}
For water segmentation, we apply Deeplab v3+\cite{chen2018encoder}, which is an effective model for semantic segmentation tasks. We evaluate the model on the high- and low-resolution images separately in our water segmentation benchmark. We also evaluate the model independently on the images of both resolutions. The training set and test set contain images captured in different scenes and under different weather conditions. For evaluation, we refer to the evaluation metrics of the lane detection task \cite{fritsch2013new}. The number of images in the training set and test set as well as the test results are shown in Table.~\ref{table: water segmentation Deeplab v3+}.
%test the generalization ability of the model and 
% MaxF: Maximum F1-measure, AP: Average precision as used in PASCAL VOC challenges, PRE: Precision, REC: Recall, FPR: False Positive Rate, FNR: False Negative Rate

\begin{figure}[t]
\centering
% \vspace{0.1in}
% \includegraphics[width=0.15\textwidth]{picture/water_seg/result/good Res/N03_5_0000018050.pdf}
% \hspace*{-0.08in}
% \includegraphics[width=0.15\textwidth]{picture/water_seg/result/good Res/W06_3_0000007850.pdf}
% \hspace*{-0.08in}
% \includegraphics[width=0.15\textwidth]{picture/water_seg/result/good Res/X31_1_0000016700.pdf} 
% \vspace*{0.01in}
\includegraphics[width=0.156\textwidth]{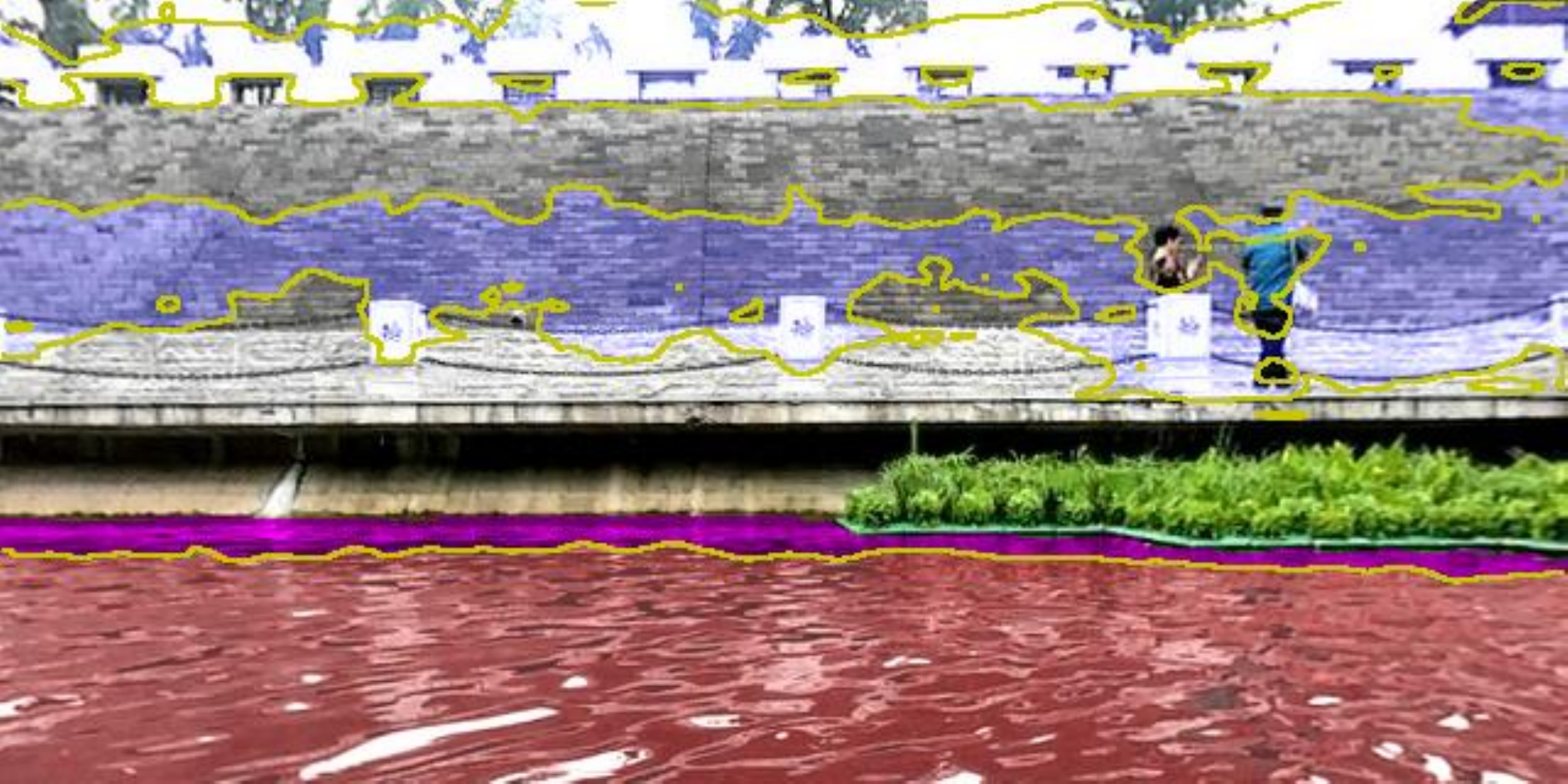}
\hspace*{-0.08in}
\includegraphics[width=0.156\textwidth]{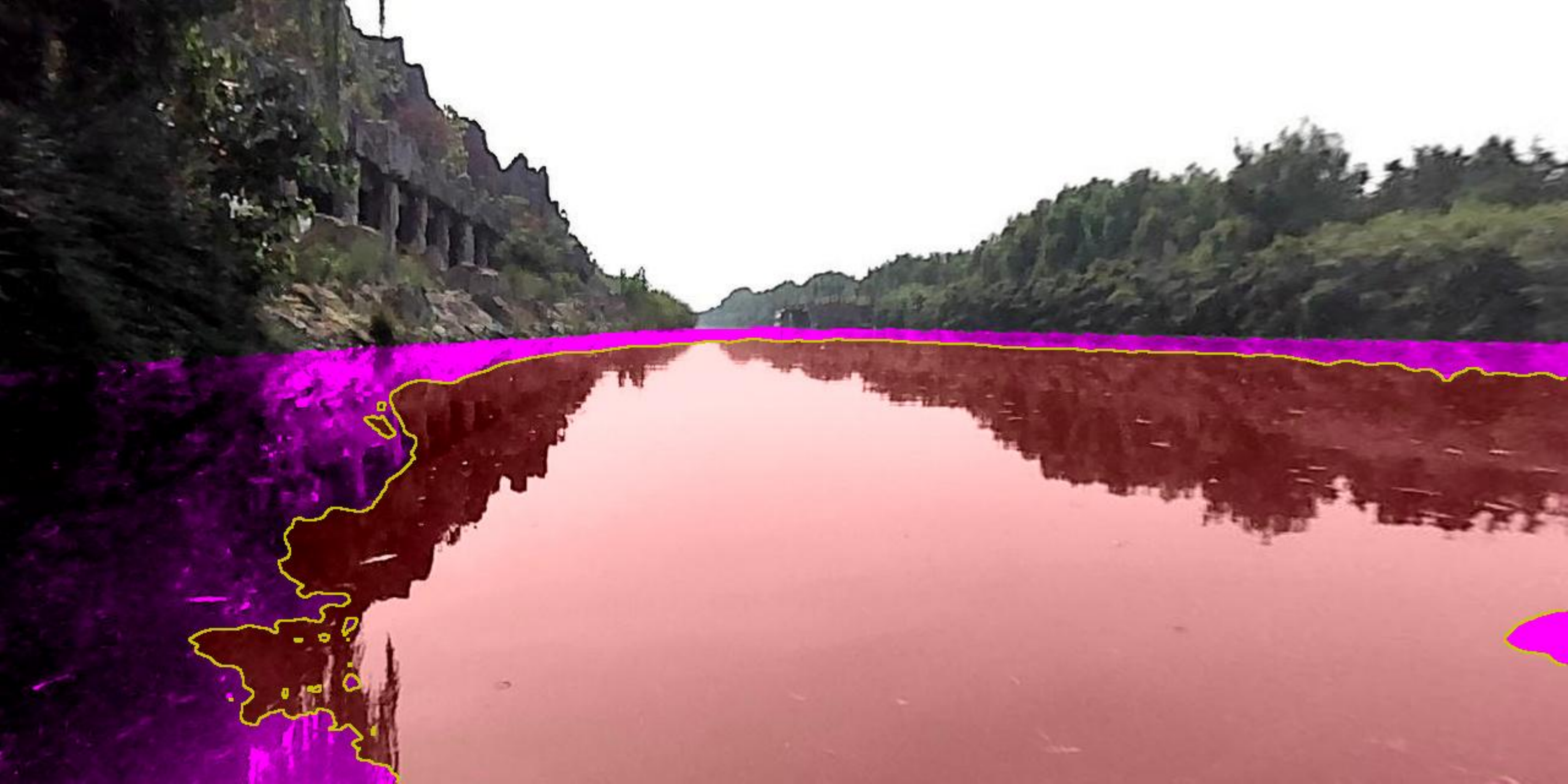} 
\hspace*{-0.08in}
\includegraphics[width=0.156\textwidth]{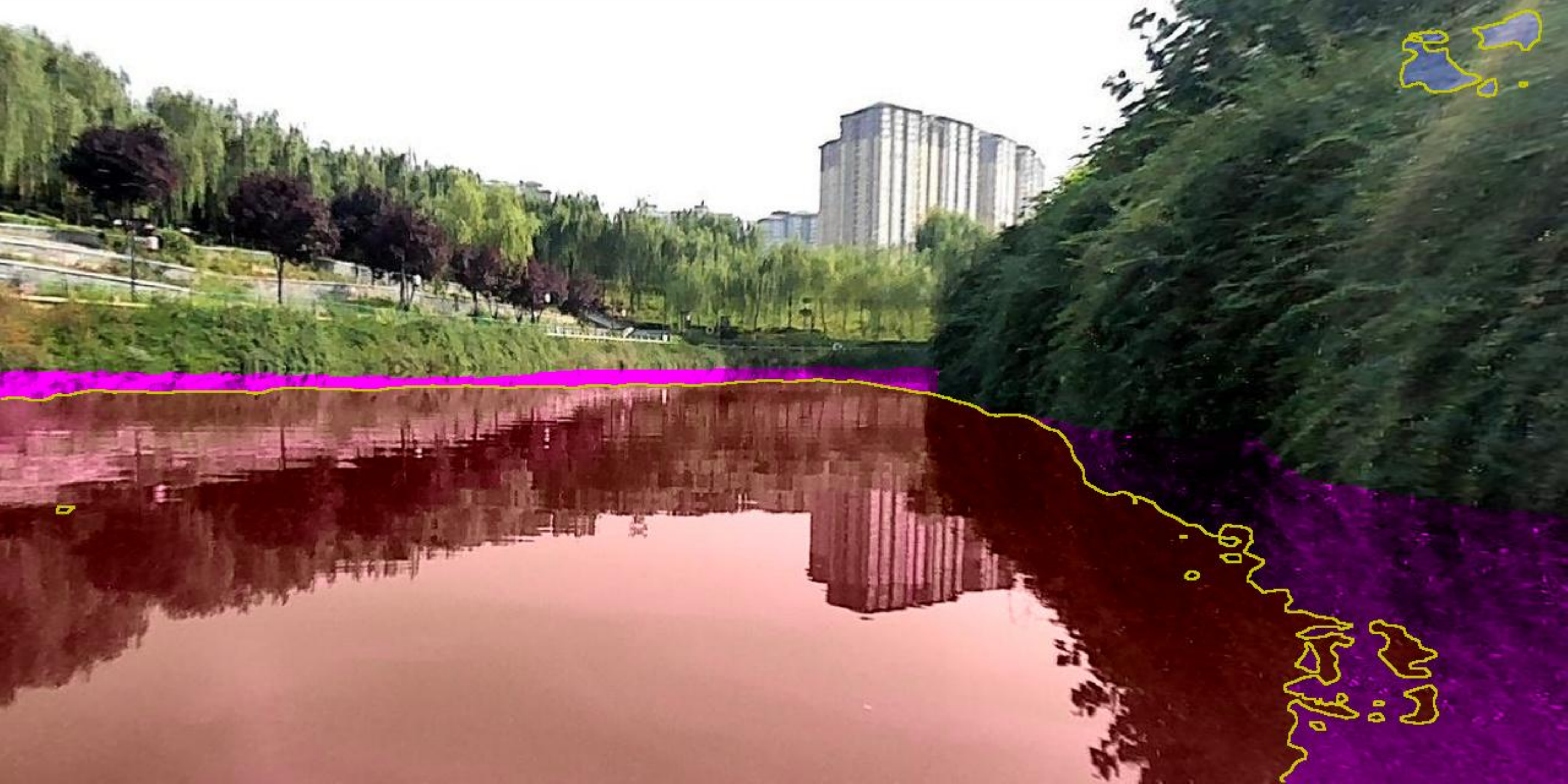} 

\caption{Examples of the water segmentation results based on DeepLab v3+: Red area: true positive. Purple area: false negative. Blue area: false positive. The yellow line shows the estimated waterline. The image on the left shows that, for one image in the test set, the wall on the bank is falsely detected as water. The other two images show that it is difficult to accurately segment the water area especially for areas near the water boundaries.}
\label{fig:Segmenation Result}
\end{figure}

We present images with low-accuracy results in Fig.~\ref{fig:Segmenation Result} to illustrate the difficulties in the water segmentation task. 
The high degree of variability of different inland water environments makes it difficult to build a robust model for water segmentation \cite{scherer2012river}. Also, unlike lane detection tasks for autonomous vehicles, the similarity between the appearance of objects on the banks and the reflection makes it difficult to accurately segment the water area. 
%It is hoped that there will be solutions for more accurate water segmentation in the future.
% Misclassification also happens in the water area with distinct wave textures and in rain. 
% The model has learnt features of water areas but still not robust to the changes of scenes.
% We hope that the generalization ability of the model be well improved in the future To increase the accuracy,. 
% Percentage of erroneous pixels in total
% result of deeplab v3+ on kitti
%-----94.38 %	92.72 %	94.70 %	94.06 %	2.40 %	5.94 %

%%%%%%%%%%%%%%%%%%%%%%%%%%%%%%%%%%%%%%%% figure_stereo %%%%%%%%%%%%%%%%%%%%%%%%%%%%%%%%%%%%%

\begin{table}[t]
\vspace{0.1in}
\caption{Result of stereo matching based on SGM}
% \centering
\begin{center}
\begin{tabular}{c|c|c|c|c|c}
\hline
{Threshold} & {5px} & {4px} & {3px} & {2px} & {1px}\\
\hline
% {Accuracy\tnote{*}}&{93.120}&{90.714}&{87.435}&{80.858}&{51.197} \\
% \hline
Out-All\tnote{*}&{6.880}&{9.286}&{12.565}&{19.142}&{48.803} \\
\hline
\end{tabular}
\end{center}
\begin{tablenotes}
\footnotesize
\item[*] Out-All: Percentage of erroneous pixels in total
\end{tablenotes}
\label{table:Stereo Matching SGM}
\end{table}

%%%%%%%%%%%%%%%%%%%%%%%%%%%%%%%%%%%%%%%% figure_stereo %%%%%%%%%%%%%%%%%%%%%%%%%%%%%%%%%%%%%
%%%%%%%%%%%%%%%%%%%%%%%%%%%%%%%%%%%%%%%% figure_stereo %%%%%%%%%%%%%%%%%%%%%%%%%%%%%%%%%%%%%

\begin{table}[t]
\vspace{0.1in}
\caption{Result of water segmentation on DeepLab v3+\tnote{*}}
\begin{center}
\begin{tabular}{c||c|c|c}
\hline
{Resolution}&{\textbf{Low-Res}}&{\textbf{High-Res}}&{\textbf{Both}}\\
\hline
{Training Num}&{345}&{121}&{466}\\
% \hline
{Test Num}&{173}&{61}&{234}\\
% \hline
{\textbf{MaxF}[\%]}&{94.52}&{96.02}&{95.80}\\
% \hline
{\textbf{AvgPrec}[\%]}&{94.12}&{95.13}&{94.99}\\
% \hline
{\textbf{PRE}[\%]}&{99.11}&{99.55}&{98.93}\\
% \hline
{\textbf{REC}[\%]}&{90.33}&{92.32}&{93.28}\\
% \hline
{\textbf{FPR}[\%]}&{0.65}&{0.43}&{1.27}\\
% \hline
{\textbf{FNR}[\%]}&{9.67}&{7.68}&{6.72}\\
\hline
\end{tabular}
\label{table: water segmentation Deeplab v3+}
\end{center}

% \begin{tablenotes}
% \footnotesize
% \item[*] Note: 
% The meaning of the evaluation metric can be found in \cite{fritsch2013new}.
% \end{tablenotes}
\end{table}

%%%%%%%%%%%%%%%%%%%%%%%%%%%%%%%%%% water segmentation
%%%%%%%%%%%%%%%%%%%%%%%%%%%%%%%%%% water segmentation
%%%%%%%%%%%%%%%%%%%%%%%%%%%%%%%%%% water segmentation
%%%%%%%%%%%%%%%%%%%%%%%%%%%%%%%%%% water segmentation

\section{Conclusion}
We present the USVInland dataset, the first multisensor USV dataset for inland waterways. The dataset is collected under various weather conditions to represent the real-world driving scenes in inland waterways. Based on our dataset, we build benchmarks for SLAM, stereo matching and water segmentation tasks. After evaluating typical algorithms on our dataset, we find that common approaches for autonomous driving may not perform well when applied to inland waterway scenes and do not meet the requirements for the safe navigation of USVs. By publishing the USVInland dataset, we hope that this work will support researchers interested in USVs for inland waterways and spur advancement in this field. 
% \textcolor{blue}{
In the future, we plan to further extend our dataset. For example, historical data will also be included to enable research on the long-term SLAM of USVs in real-world inland waterways.
% }

\end{document}